\documentclass[journal]{IEEEtran}
\usepackage{amssymb}
\usepackage{subfig}
\usepackage{framed,multirow}

\usepackage{latexsym}
\usepackage{amsmath}
\ifCLASSINFOpdf
   \usepackage[pdftex]{graphicx}
   \graphicspath{{images/}}
  \DeclareGraphicsExtensions{.pdf,.jpeg,.png}
\else
\fi

\usepackage{booktabs , array}

\usepackage{amsmath}

\begin{document}
\title{Pointwise Attention-Based Atrous Convolutional Neural Network}

\author{\IEEEauthorblockN{Mobina Mahdavi\textsuperscript{*},
Fahimeh Fooladgar\textsuperscript{*}\thanks{\noindent\textsuperscript{*}These authors contributed equally} and Shohreh Kasaei\textsuperscript{**}\thanks{\noindent\textsuperscript{**}Corresponding author: 
	Tel.: +98-21-6616-6646,  
	fax: +98-21-6601-9246, Email: kasaei@sharif.edu}}

\IEEEauthorblockA{Department of Computer Engineering,
Sharif University of Technology\\
Tehran, Iran\\}}


% \markboth{Journal of \LaTeX\ Class Files,~Vol.~14, No.~8, August~2015}%
% {Shell \MakeLowercase{\textit{et al.}}: Bare Demo of IEEEtran.cls for IEEE Journals}

\maketitle

\begin{abstract}
With the rapid progress of deep convolutional neural networks, in almost all robotic applications, the availability of 3D point clouds  improves the accuracy of 3D semantic segmentation methods. 
Rendering of these irregular, unstructured, and unordered 3D points to 2D images from multiple viewpoints imposes some issues such as loss of information due to 3D to 2D projection, discretizing artifacts, and high computational costs. To efficiently deal with a large number of points and incorporate more context of each point, a pointwise attention-based atrous convolutional neural network architecture is proposed. 
It focuses on salient 3D feature points among all feature maps while considering outstanding contextual information via spatial channel-wise attention modules. The proposed model has been evaluated on the two most important 3D point cloud datasets for the 3D semantic segmentation task. It achieves a reasonable performance compared to state-of-the-art models in terms of accuracy, with a much smaller number of parameters.
\end{abstract}


% \fi
\IEEEpeerreviewmaketitle

\section{Introduction}
\label{sec1}
Semantic segmentation refers to partitioning the image into meaningful segments by assigning a label to each pixel. It is one of the most important processes in the field of Computer Vision. 
It is crucially important in many applications; such as autonomous driving or robot navigation, where the intelligent agent's decision has to make sense based on the input image to be able to operate correctly.
With the emergence of methods based on deep Convolutional Neural Networks (CNNs) \cite{simonyan2014very,he2016identity,krizhevsky2012imagenet,hu2018squeeze}, prediction accuracy has increased significantly, compared to learning methods based on hand-crafted features. 

The availability of RGB-Depth cameras has changed the perspective of ongoing research. Many papers \cite{cadena2013semantic,muller2014learning,silberman2011indoor,fooladgar20193m2rnet,fooladgardicta,fooladgar2015learning} have focused on fusing the information of per-pixel depth data with the RGB data. The depth information takes into account the fact that images are not just 2D matrices, but a projection of a 3D scene.
However, the depth data makes the accurate 3D reconstruction of the scene feasible and simple. 
Therefore, many have started to explore the application of deep learning directly on the 3D data \cite{qi2017pointnet,qi2017pointnet++}.

The 3D scene can be easily represented as a point cloud, which requires much less memory as opposed to a voxel representation. It also does not suffer from the quantization error which is a consequence of voxel representations \cite{qi2017pointnet}. Point clouds are lists of point coordinates that are invariant to permutation. Hence, the extension of the state-of-the-art techniques on 2D images to 3D point clouds is non-trivial.
Utilizing the pointwise convolution \cite{hua2018pointwise} as the convolutional operator is one of the reported methods to handle this issue. In that operator, a point's nearest Euclidean neighbors contribute to its convolution. Compared to convolution on 2D, which operates on the neighboring pixels that may not be the true 3D neighbors of that point, pointwise convolution works with real neighbors. This means that the model is a few steps ahead, by knowing which points are not relevant from the beginning.

\begin{figure*}[!t]
\centering
\includegraphics[width=13cm]{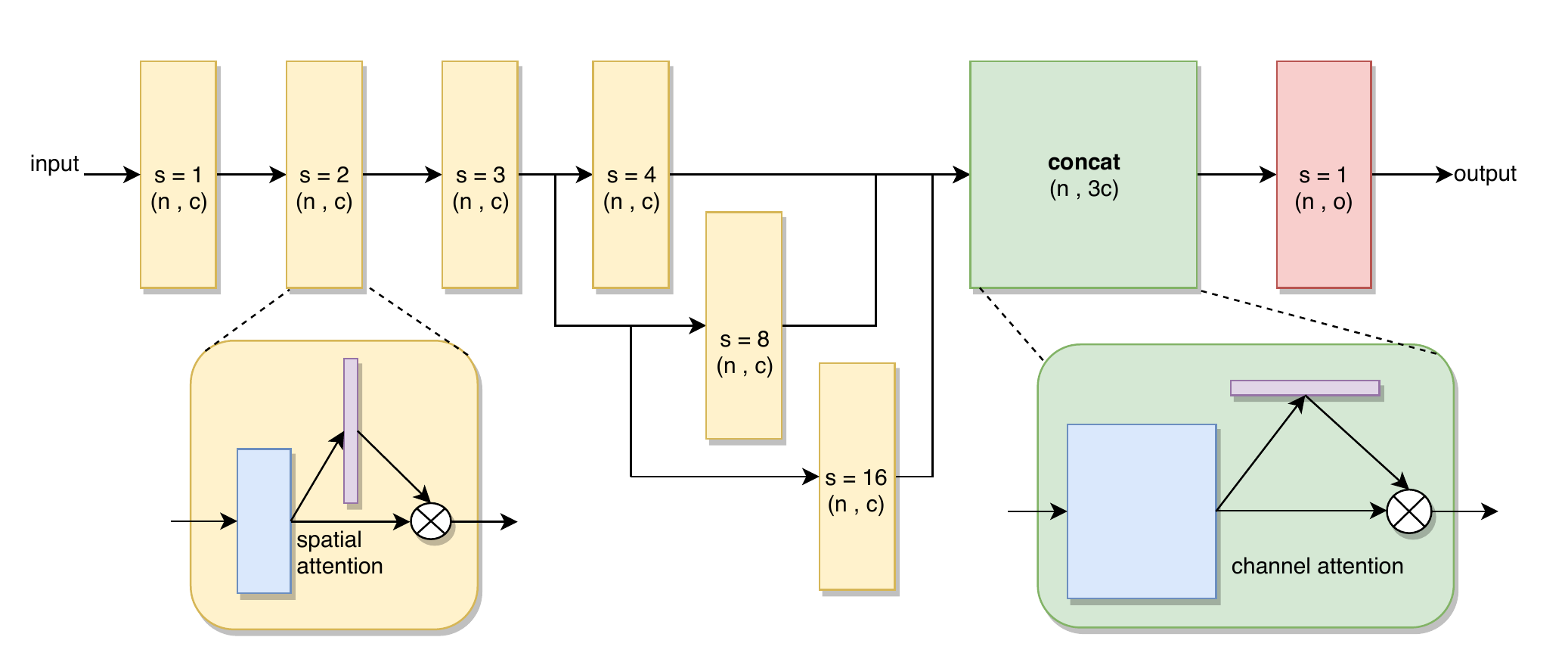}
\caption{Diagram of proposed PAAConvNet. [Feature map shape and stride of atrous convolution $s$ is written on each block. Here, $n$ is the size of a partial point cloud, $c$ is the number of input channels, and $o$ is the number of output classes. Yellow blocks are pointwise convolutional layers with spatial attention module. The green block is a concatenation layer with a channel attention module. The red block is a pointwise convolution module that outputs the labels of each point.]}
\label{fig:Diagram}
\end{figure*}

In this paper, a novel network structure is proposed. It also provides the model with additional features which yields much better results. The 3D point cloud of each area contains a large number of 3D points in comparison to each 2D image; where each 3D point cloud may be reconstructed from multiple images from multiple viewpoints. Therefore, incorporating context besides focusing the attention on salient points can yield significant improvements via the network architecture (see Figure \ref{fig:Diagram}). Hence, this work proposes to leverage the atrous convolution to incorporate more context in 3D point sets. An atrous convolution enables the increase of the receptive field while preserving the same number of parameters and hence the computational cost. It has shown promising results \cite{chen2018deeplab_j,chen2016deeplab}. The proposed model uses increasingly bigger strides in cascade and parallel manners which helps the model to detect objects from multiple sizes. 
In the proposed method, to find the more salient 3D points and feature maps, the attention modules have been proposed. The proposed spatial- and channel-wise attention modules are learned and multiplied into pointwise convolutional layers to impose the attention of weights into important points or feature vectors, and thus help the model learn significantly faster. Moreover, point clouds only contain the coordinates of points with their RGB colors. Of course, additional 3D hand-crafted features can be computed and exploited as an additional input. It is proposed to incorporate 3D surface normal alongside 3D point coordinates and RGB colors as initial feature vectors. 

The main contributions of this work are listed as:
\begin{itemize}
	\item Proposing the pointwise atrous convolution for 3D point cloud semantic segmentation.
	\item Incorporating an attention mechanism of CNNs to determine the salient 3D points as well as salient feature maps.
\end{itemize}

The remainder of this paper is organized as follows. In Section \ref{sec:2}, the related semantic segmentation methods are categorized. The overall architecture of the proposed Pointwise Attention-Based Atrous Convolutional Neural Network (PAAConvNet)  is presented in Section \ref{sec:3}. The experimental results evaluated on the existing 3D point cloud datasets are investigated in Section \ref{sec:4}. Finally, concluding remarks are presented in Section \ref{sec:5}.    

%%%%%%%%%%%%%%%%%%%%%%%%%%%%%%%%%%%%%%%
%%%%%%%%%%%%%%%%%%%%%%%%%%%%%%%%%%%%%%%
\section{Related Work}
\label{sec:2}
Semantic segmentation is a wide-spreading research area. There are many outstanding methods presented for RGB and RGB-Depth images, from traditional hand-crafted features \cite{shotton2006textonboost,yang2011hierarchical,zheng2014dense,zhu2016beyond} to recent deep convolutional neural networks \cite{long2015fully,chen2016deeplab,chen2018deeplab_j,chen2018encoder,cadena2013semantic,fooladgar2015learning,fooladgardicta,muller2014learning,silberman2011indoor,fooladgar20193m2rnet}. 

The goal of 3D point cloud semantic segmentation is to partition the 3D points to semantically coherent 3D regions where a label is assigned to each point. Hence, it deals with reasoning about 3D geometric data. Incorporating common CNNs \cite{he2016identity,hu2018squeeze,huang2017densely,krizhevsky2012imagenet,simonyan2014very,szegedy2015going} on 3D point sets which are irregular and non-uniform structures produces inappropriate results. 

Traditional approaches for semantic segmentation utilize engineering features to classify each pixel, superpixel, 3D point, or voxel. The hand-crafted features extracted from 3D points have been exploited for various tasks such as shape analysis \cite{aubry2011wave,bronstein2010scale}, 3D point cloud registration \cite{rusu2009fast}, 3D model retrieval \cite{chen2003visual}, and shape classification \cite{ling2007shape}. They are designed to be invariant to some transformations such as pose, sampling density, rotation, and translation. 

The spatially local features of convolution operations are incorporated in CNN architectures. They demonstrate remarkable improvements in various applications on regular data domains such as images and videos. But, this significant property of convolution operation cannot be effectively applied to irregular and unordered 3D point clouds. To utilize CNN models for 3D point clouds and 3D meshes, four types of input structures have been reported in the literature. 
\begin{itemize}
	\item  The simplest way is to project each point cloud on 2D images from multiple viewpoints \cite{tatarchenko2018tangent, lawin2017deep,su2015multi,boulch2016deep}. This 3D to 2D projection imposes information loss, which is diminished by using multiple viewpoints.   

 \item The low-level local and global geometric features can be extracted from 3D point sets and then fed to a convolutional model \cite{boulch2016deep,guo20153d,rusu2009fast,guo2013rotational,marton2011combined}. The performance of these approaches depends on the quality of the extracted features.

\item The regular 3D volumetric blocks of irregular point clouds, called voxels, have been considered as inputs to deep neural network models \cite{liu20173dcnn, wang2017cnn,riegler2017octnet}. There exists a trade-off between the computational cost and the quantization error.

\item In the fourth category, 3D point clouds are directly fed to deep neural networks to learn their structure via some transformed mini-networks or special convolution operations \cite{qi2017pointnet,hua2018pointwise,wang2018dynamic}. 
\end{itemize}

In \cite{ben20183dmfv}, the 3D Modified Fisher Vector (3DmFV) is proposed for 3D point cloud presentation where each point is described by its deviation from the Gaussian Mixture Model (GMM). In this representation, the discrete nature of 3D point clouds is combined with the continuous characteristics of GMM.  
Recently, the improvement of convolutional neural network architectures for 3D meshes and 3D geometric data points has concentrated more attention \cite{qi20173d,qi2017pointnet,qi2017pointnet++,hua2018pointwise}. 
The PointNet \cite{qi2017pointnet} utilizes an unordered set of points and learns transformation functions via CNNs to be invariant to rotation and translation of entire points of each point cloud. The PointNet++ \cite{qi2017pointnet++}, an extension of the PointNet, leverages the hierarchical feature learning method to improve the accuracy. 

In \cite{hua2018pointwise}, the idea of pointwise convolution operation is proposed to deal with unordered and irregular 3D point clouds. Therefore, every convolution operation is applied on each 3D point based on its Euclidean neighboring points. In \cite{li2018pointcnn}, the idea of ordered-equivariance $\chi$-transformation is presented. In \cite{thomas2019kpconv}, rigid and also deformable Kernel Point Convolution (KPConv) operations are used which directly apply on point clouds. In \cite{xiong2019deformable}, 3D deformable filters are applied on 3D point clouds (instead of 3D discretizing via voxel) to preserve the model against loss of local geometric information.

%%%%%%%%%%%%%%%%%%%%%%%%%%%%%%%%%%%%%%%
\section{Proposed Pointwise Attention-Based Atrous Convolutional Neural Network Method}
\label{sec:3}
The proposed PAAconvNet method takes a partial point cloud as the input and classifies each point. In this work, first, the partial point cloud is augmented with additional features and then fed into the PAAconvNet.  

Each point cloud has lots of 3D points where the overall shape of points is unstructured and unordered. Therefore, utilizing pointwise convolution is more promising than the common convolution operator, for well incorporating the spatially local information. Besides, among these large numbers of points, focusing on salient 3D points or features can significantly improve the labeling performance. Also, the contextual relations among these points can play an important role in semantic labeling. To incorporate the context on irregular salient point sets, the attention-based atrous pointwise convolution operation is proposed. In the following, the method is explained in detail.

\subsection{Atrous Pointwise Convolution}

The PAAConvNet mainly consists of convolutional layers. Due to the unstructured characteristic of point clouds, a special kind of convolutional operator (pointwise convolution \cite{hua2018pointwise}) is utilized. In every such layer, the partial point cloud is pre-processed and every point is put in a cell of a 3D grid. In general, the convolution of point $X$ can be computed as

\begin{equation}
	conv(X) = \sum_{k}w_k \bar{x_k}	 \label{eq:conv}
\end{equation}

where $k$ is the index of a cell of the convolutional kernel, $w_k$ is the weight associated with it, and $\bar{x_k}$ is the mean of points within that cell (which is zero if the cell is empty). Note that based on the shape of $w_k$, $X$ can have a different number of channels than $conv(X)$. Finally, an activation function is applied over the convolution function.  

In this formulation, the kernel can have any shape. Therefore, in this work, the atrous convolutional kernel is defined as a $3\times3\times3$ cube, with a weight of $W_{(i , j , k)}$ associated with each of its unit cubes, where $i,j,k$ are either $0$, $1$, or $-1$. Given that point $X$ is located at cell $C_{(x , y , z)}$, an atrous convolution of $X$ with stride $s$ can be written as

\begin{equation}
	conv(X) = \sum_{-1\leq i,j,k\leq 1} W_{(i , j , k)} \bar{C}_{(x + si , y +sj , z+sk)}	 \label{eq:conv_atrous}
\end{equation}

where $\bar{C}_{x , y , z}$ is the mean of feature points within that cell.

One of the important hyper-parameters of the model is the size of the cells. If it is too large, a larger area is considered for the convolution of a single point. As such, a lot of neighboring points will contribute to that cell. Hence sharing the same weight that applies to the average of them makes the model less sensitive to smaller local features. Also, smaller details would be smoothed out.

Thus, to have a greater field of view, it is better to increase the kernel size rather than the cell size (utilizing a bigger cube; e.g., $5\times5\times5$). But, this approach is computationally expensive. One of the best alternatives to this scheme is the utilization of atrous convolution that helps to increase the field of view with just a little additional cost.
 
In PAAConvNet, the kernel size is fixed ($3\times3\times3$) and different strides for atrous convolution are used; with a fine to coarse approach. In the early stages of the network, the stride starts from one (common convolution) and subtly increases for each layer in the cascade, to detect fine details.
On the contrary, near the end stages of the network, exponentially increasing strides are used in parallel to detect coarser ones. Then, the extracted feature maps are concatenated. The details of PAAConvNet can be seen in Figure. \ref{fig:Diagram}. 

Note that throughout the model, no kind of down-sampling is applied to lower the resolution of feature maps in the partial point cloud. Therefore, the resolution of output points stays the same as input points and the model incorporates the contextual information through the atrous convolution without any down-sampling operation.

\subsection{Attention Module}

\begin{figure}[!t]
\centering
\includegraphics[width=6cm]{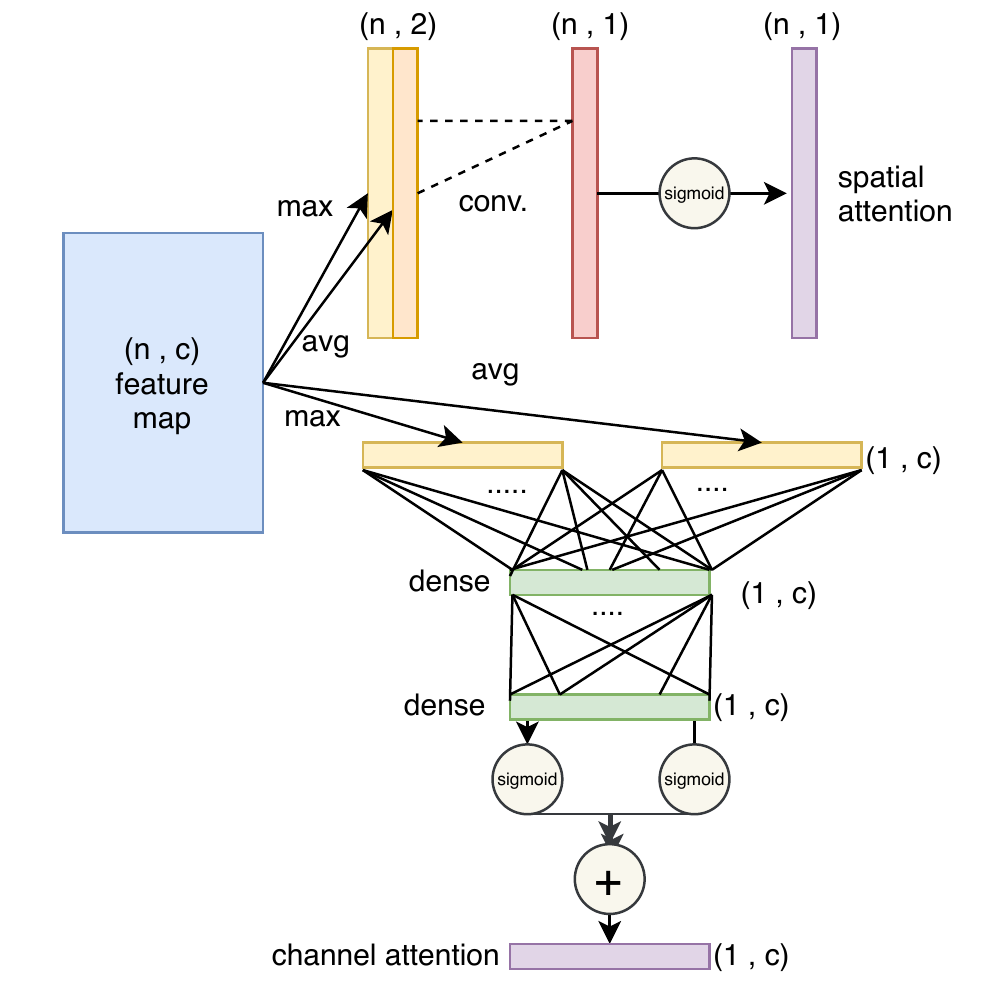}
\caption{Details of obtaining spatial and channel attention modules, from a feature map of $n$ points with $c$ channels.}
\label{attention_fig}
\end{figure}

The convolutional attention modules bring the attention of the model to both spatially significant points in the partial point cloud or more important feature maps. Spatial attention and channel attention modules can be added to each layer. As shown in Figure.  \ref{attention_fig}, they are computed and learned through the back-propagation alongside other parameters in the network. 

In each layer, the feature map is $F \in \mathbb{R}^{n\times c}$, where $n$ is the number of points and $c$ is the number of channels. The spatial attention module is $S \in \mathbb{R}^{n\times 1}$, which is multiplied into every column of $F$. In order to obtain $S$, we have
\begin{equation}
	S = \sigma(conv1D([F_{max}^n;F_{avg}^n] , f))
\end{equation}

where $F_{max}^n  , F_{avg}^n\in \mathbb{R}^{n\times1}$ are the results of max pooling and average pooling of $F$ over its second dimension that are concatenated to form a $n\times2$ feature map, $f$ is a kernel with size of 5, and $\sigma$ is the sigmoid function.

 It is important to note that in this model, the input partial point cloud is sorted. Therefore, the sequence of points is consistent for all data and the one-dimensional convolution is appropriate. 

The channel attention module is $C \in \mathbb{R}^{1\times c}$, which is multiplied into every row of $F$. For obtaining $C$, we have
\begin{equation}
	C = \sigma(dense(F_{avg}^c) + dense(F_{max}^c))
\end{equation}
where $F_{max}^c  , F_{avg}^c\in \mathbb{R}^{1\times c}$ are results of max pooling and average pooling of $F$ over its first dimension, $dense$ is a neural network with two layers of size $c$ that is shared among them, and $\sigma$ is the sigmoid function.

In the early blocks of PAAConvNet, before the concatenation layer, the spatial attention modules are applied to focus on salient 3D points. After this concatenation layer, the number of channels increases threefold, hence a channel attention module is added to determine the salient channel-wise features.  

\subsection{PAAConvNet Loss Function}
In this model, each partial point cloud can be defined as 

\begin{equation}
\label{eq:2}
\begin{split}
\{(X_{i},RGB_{i},N_{i},Y_i ) |  X_{i} ,  RGB_{i} , N_i \in \mathbb{R}^{(1\times 3)}, \\ Y_i  \in L, i=1,…,N_c\}
\end{split}
\end{equation}
where  $X_{i} , RGB_i , N_i, Y_i$ denote the 3D coordinates, RGB color, surface normal, and category label of each point $i$, respectively,
$N_c$ determines the number of points in each partial point cloud, and $L$ denotes the labeling set defined as
$L= \{ 1,…,C \} $ 
in which $C$ determines the number of class labels. 

The simple pointwise categorical cross-entropy for the loss function is defined as
\begin{equation}
\label{eq:3}
\ell(y_i,\hat{y_i}) = -\sum_{c=1}^{C} y_{i,c} \log p(\hat{y_i}=c | PC_i ,\mathbb{W})
\end{equation}
where $PC_i=\{X_i,RGB_i,N_i\}$ and $y_{i,c}$ is the one-hot encoding vector of ground-truth label of each point $i$ with $C$ labels. The probability that the point $PC_i$ predicts class $c$ is computed by
\begin{equation}
\label{eq:4}
p(\hat{y_i}=c |  PC_i  ,\mathbb{W})=\frac{e^{(f_c (PC_i;\mathbb{W}))}}{\sum_{l=1}^{C} e^{(f_l (PC_i;\mathbb{W}))}}.
\end{equation}
where $f_c (PC_i;\mathbb{W})$ is the output of PAAConvNet before the softmax layer and $\mathbb{W}$ denotes all of  network parameters.

\begin{table}[!t]
\caption{Comparison of proposed PAAConvNet and its variants with PointWise model over mean class accuracy (mAcc) and overall accuracy, on S3DIS dataset.}
\label{s3dis}
\centering
\begin{tabular}{|c|cc|}
\hline
Model & mAcc(\%) &overall accuracy(\%)\\
\hline
PointWise & 56.5 & 81.5\\
base model(c = 9)(ours) & 63.7& 84.7\\
base model(c = 12)(ours) & 67.7 & 86.2\\
Proposed PAAConvNet & \textbf{74.2}\% &\textbf{87.3}\%\\

\hline
\end{tabular}
\end{table}

\begin{table*}[!t]
\caption{Comparison of 6-fold overall accuracy (OA) of S3DIS, with accuracy of every class compared to PointNet\cite{qi2017pointnet}, SGPN\cite{wang2018sgpn}, G+RCU\cite{3dsemseg_ICCVW17}, SCN\cite{Xie_2018_CVPR},
 SPG\cite{DBLP:journals/corr/abs-1711-09869},DGCNN\cite{dgcnn},  PointCNN\cite{li2018pointcnn}, LAE-Conv\cite{feng2019point}, and RSNet\cite{huang2018recurrent}  .}
\label{kfold_s3dis}
\resizebox{\textwidth}{!}{
\centering
\begin{tabular}{|c|c|ccccccccccccc|}
\hline
Model & OA (\%) &ceiling&	floor&	wall&	beam	&column&	window&	door&	chair	&table&	bookcase&	sofa&	board&	clutter
\\
\hline
PointNet & 78.5 & 88&	88.7&	69.3&	42.4&	23.1&	47.5&	51.6&	54.1&	42&	9.6	&38.2&	29.4&	35.2
 \\
SGPN &80.8  &-&-&-&-&-&-&-&-&-&-&-&-&-\\
G+RCU &81.1&-&-&-&-&-&-&-&-&-&-&-&-&-\\
SCN  &81.6 &-&-&-&-&-&-&-&-&-&-&-&-&-\\

% DPRNet & 83.8 & 97.8&	99.4&	89.5&	66.9&	-&	54.9&	71.7&	71.1&	81.7&	59.5&	-&	-&	70.8
% \\

DGCNN&84.1 &-&-&-&-&-&-&-&-&-&-&-&-&- \\ %2019

SPG& 85.5 &89.9&	95.1&	76.4&	62.8&	47.1&	55.3&	68.4&	73.5	&69.2&	63.2&	45.9&	8.7	&52.9
 \\

% 3DRCNN & 85.7 & 95.2&	98.6&	77.4&	0.8&	9.83&	52.7&	27.9&	78.3&	76.8&	27.4&	58.6&	39.1&	51\\

PointCNN& 88.14 & 94.78&	97.3&	75.82&	63.25&	51.71&	58.38&	57.18&	71.63&	69.12&	39.08&	61.15&	52.19	&58.59\\

LAE-Conv & 88.95 & 94.3 & 	97 & 	76.02 & 	64.66 & 	53.7 & 	59.17	 & 58.8 & 	72.4 & 	69.2	 & 42.63 & 	60.83 & 	54.14 & 	59.05\\

RSNet & - & 92.48&	92.83&	78.56&	32.75&	34.37&	51.62&	68.11&	60.13&	59.72&	50.22&	16.42&	44.85&	52.03
\\
\hline

Proposed PAAConvNet  & {83.6} & \textbf{96.27}	&\textbf{97.83}&	\textbf{85.19}&	43.44&	40.50&	38.31&	66.22&	73.12&	\textbf{74.92}&	31.49&	55.33&	21.60&	\textbf{67.06}
\\

\hline
\end{tabular}}
\end{table*}

\begin{table*}[!t]
\caption{Comparison of per-class accuracy of some classes, on SceneNN dataset.}
\label{table:scenenn}
\centering
\begin{tabular}{|c|cccccccccc|}
\hline
Model & wall & floor & cabinet&bed &chair&sofa&table&desk&tv&prop\\
\hline

DPRNet\cite{arshad2019dprnet} & 87.4&	94.7&	22.8&	38&	70	&10	&42.8&	38.3&	26.3&	-\\
PointNet & 89.7&	89.1&	9&	45.7&	59.6&	16.7&	23.5&	31&	11.4&	-\\
SemanticFusion\cite{SemanticFusion} & 72.8&	94.4&	-&	46.3&	-&	-&	70.1&	28.1&	-&	-\\
VoxNet\cite{Maturana-2015-6018} & 82.8&	74.3&	-&	0&	3.1&	-&	0.8	&5.4&	-&	-\\
PointWise & 93.8 & 88.6 & 1.5 & 11.6 &58.6 & 5.5& 23.5&29.5&7.7&5.8\\
\hline
Proposed PAAConvNet & 84.7&87.4 & \textbf{25.1} & \textbf{51.9} & 59.3 & \textbf{17.7}& 31.5 &\textbf{45.4}&24.2& \textbf{6.7}\\

\hline
\end{tabular}
\end{table*}

\section{Experimental Results}
\label{sec:4}
The proposed method has been evaluated on two main 3D point cloud datasets. In the following subsection, more details of experimental results are elucidated.

\subsection{Evaluation Results on Stanford 2D-3D-Semantic Dataset}
The performance of PAAConvNet is examined on the S3DIS \cite{2017arXiv170201105A} dataset. It consists of point clouds of six areas in indoor scenes. Points in a one squared-meter area are sampled and grouped to blocks of 4096 as a partial point cloud. Each point belongs to one of 13 categories and has 9 channels of: coordinates regarding the block, RGB color, coordinates regarding the room.  
A batch of 4096 blocks is fed to the PAAConvNet. The network is optimized with mini-batch gradient descent with momentum.  The PAAConvNet has been trained with the first 5 areas and has been evaluated on the sixth area as a test set, to compare results with Pointwise. The evaluation metrics are the overall and per-class accuracy.

First, a base model (which is the same as Figure \ref{fig:Diagram} but with no attention module) has been evaluated to consider the effect of atrous convolution in parallel and cascade modules.

Moreover, partial point clouds in S3DIS do not include surface normals. For each point, its nearest neighbors are queried with a k-d tree search algorithm and the best plane is fit to it. Since there are two correct answers for the surface normal, these have to be consistently oriented for all of the data. The origin point is empirically chosen to be the orientation center.  Therefore, 3 more channels are added to each point which denote its surface normal. These are added in two different manners. First, they are added instead of the first XYZ coordinates, which seem redundant compared to the other set of more general coordinates. Second, they are added as extra data increasing the channel size from 9 to 12. In the second scenario the $c$ parameter in the model increases as well from 9 to 12, which leads to increased training time as well as higher accuracy. Either way, as shown in Table \ref{s3dis}, better results are produced, compared to the base model and compared to PointWise.
The increase in performance is mainly due to the exploitation of planar quality of man-made scenes, that is made available through the incorporation of surface normals. Also, having $c = 12$ makes up for more parameters in the model to be learned and utilized, which contributes to a more complex model.

\begin{figure}[!t]
\centering
\includegraphics[width=8cm]{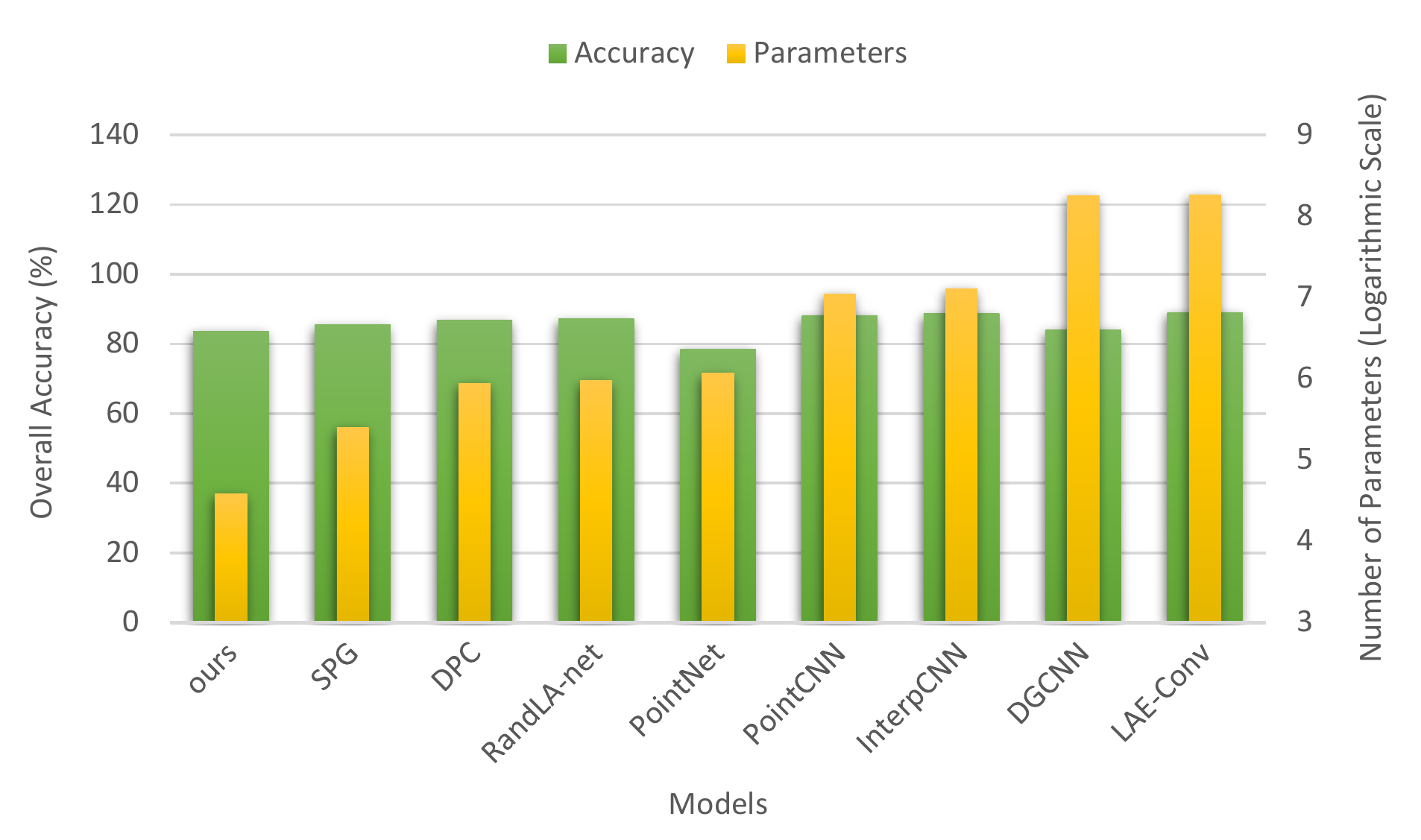}
\caption{Number of parameters (in logarithmic scale) vs. performance of proposed model compared to SPG\cite{DBLP:journals/corr/abs-1711-09869}, DPC\cite{Engelmann2019DilatedPC},
RandLA-net\cite{RANDLA}, pointnet, PointCNN\cite{li2018pointcnn}, InterpCNN\cite{Mao2019InterpolatedCN} , DGCNN\cite{dgcnn}, and LAE-Conv\cite{feng2019point}.}
\label{fig:modelsize}
\end{figure}

\begin{figure}[!t]
\centering
\includegraphics[width=7cm]{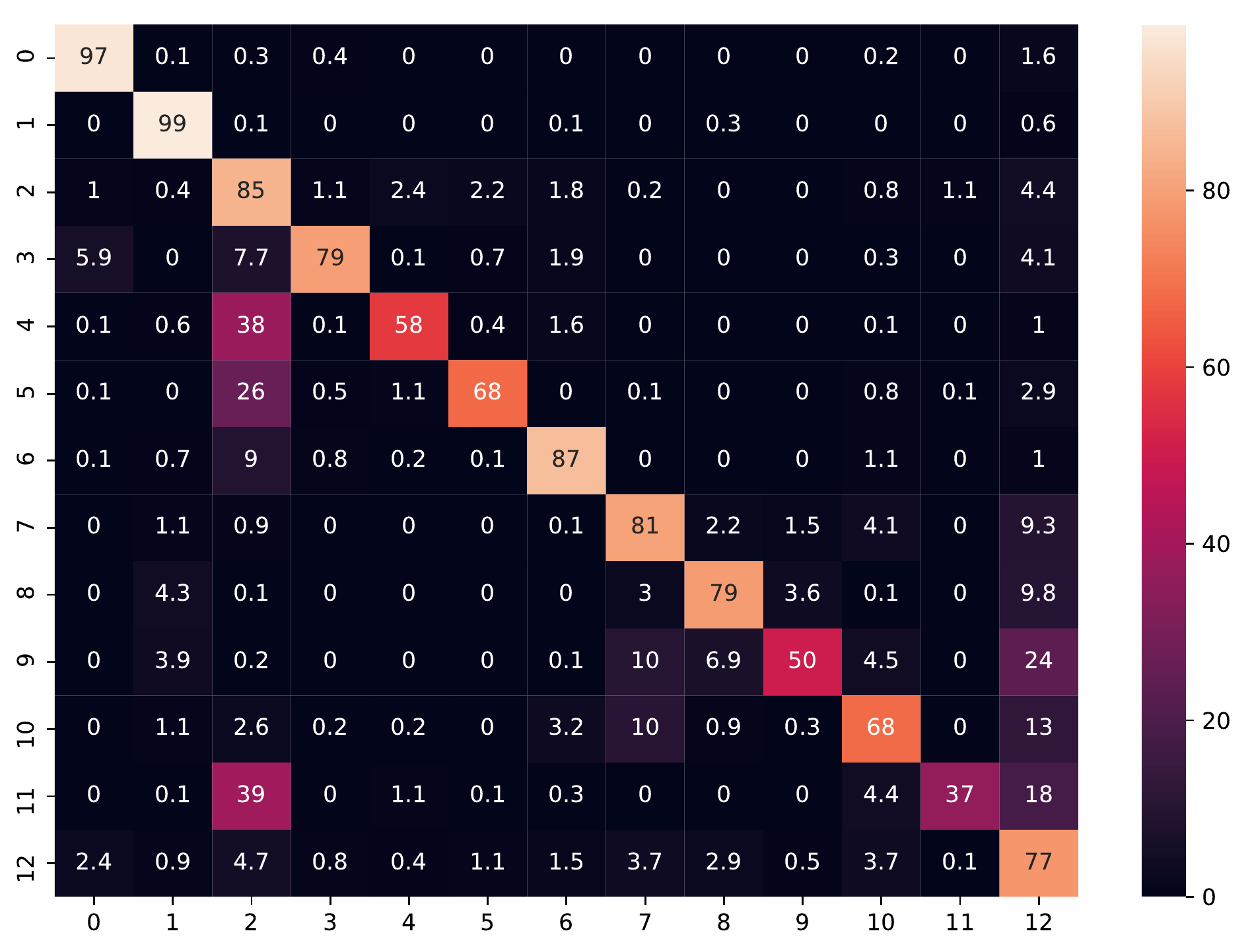}
\caption{The confusion Matrix result of testing on the S3DIS dataset, area 6. The mean IoU is $62.9\%$}
\label{fig:Confusion}
\end{figure}

\begin{figure}
\subfloat[RGB]{\hspace{2.2cm}}\subfloat[Normals]{\hspace{2.2cm}}\subfloat[GT]{\hspace{2.2cm}}\subfloat[Prediction]{\hspace{2.2cm}}
\vspace{0.1cm}
\subfloat{\includegraphics[width=0.24\linewidth]{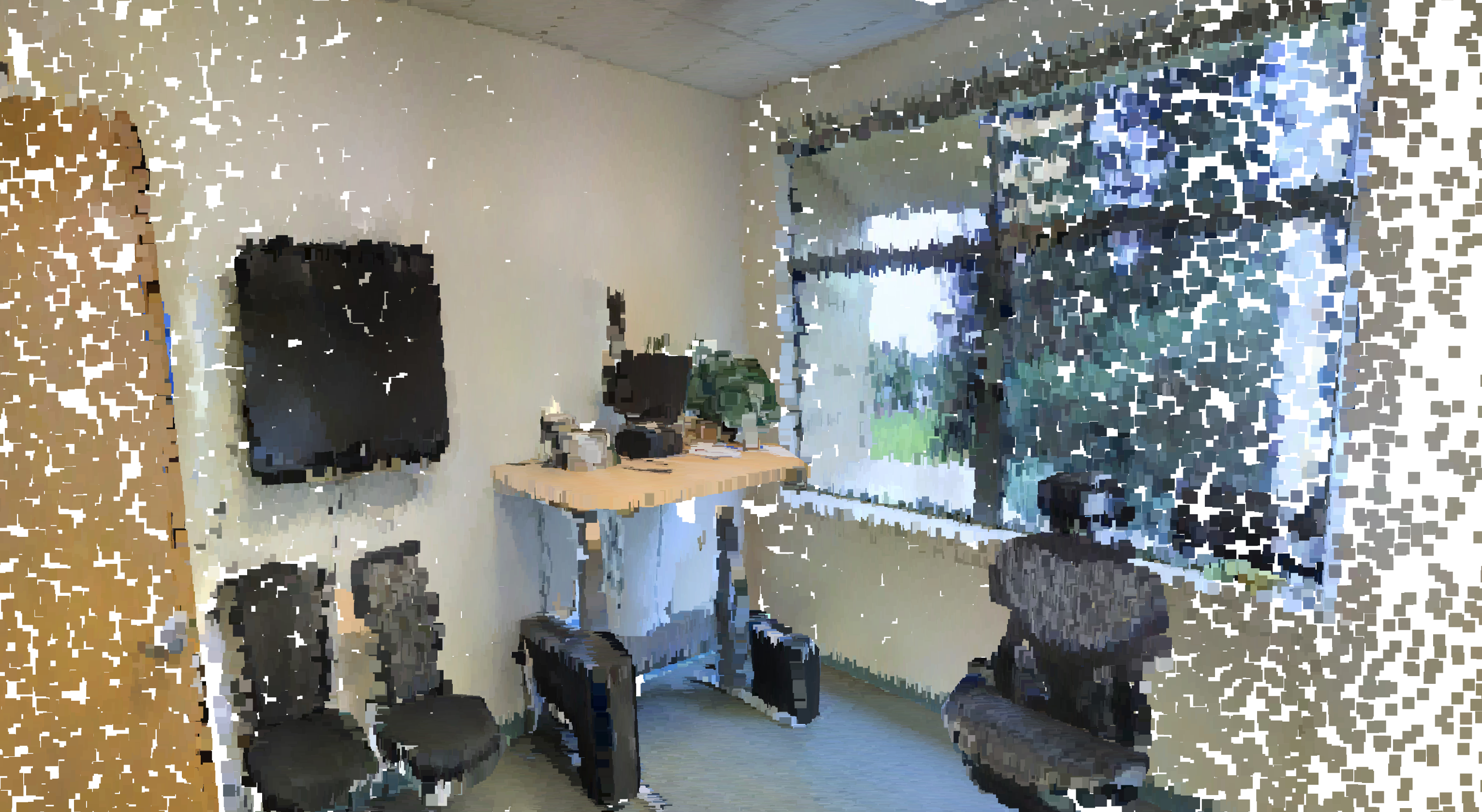}} \hspace{0.001cm}
\subfloat{\includegraphics[width=0.24\linewidth]{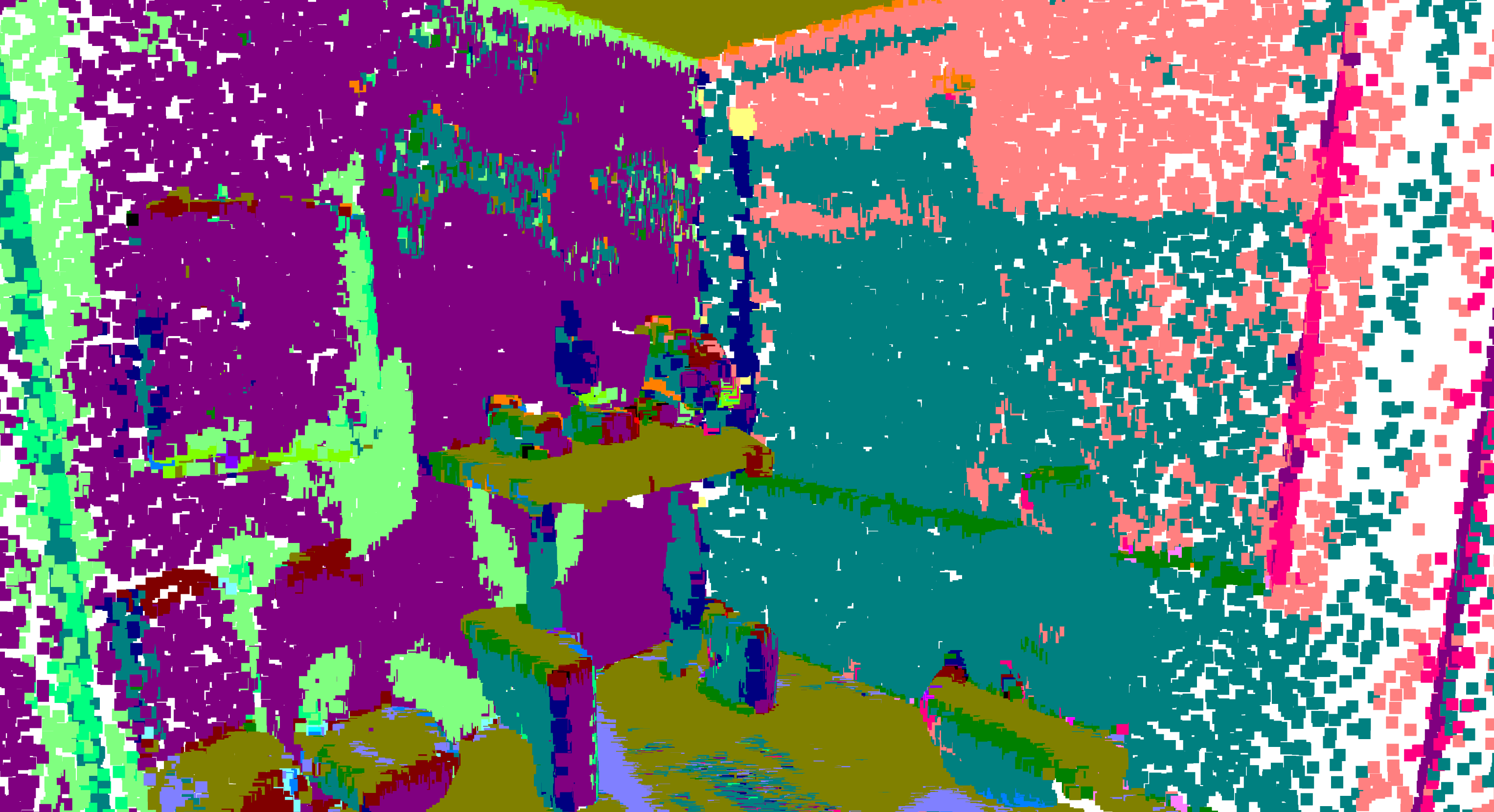}}\hspace{0.001cm}
\subfloat{\includegraphics[width=0.24\linewidth]{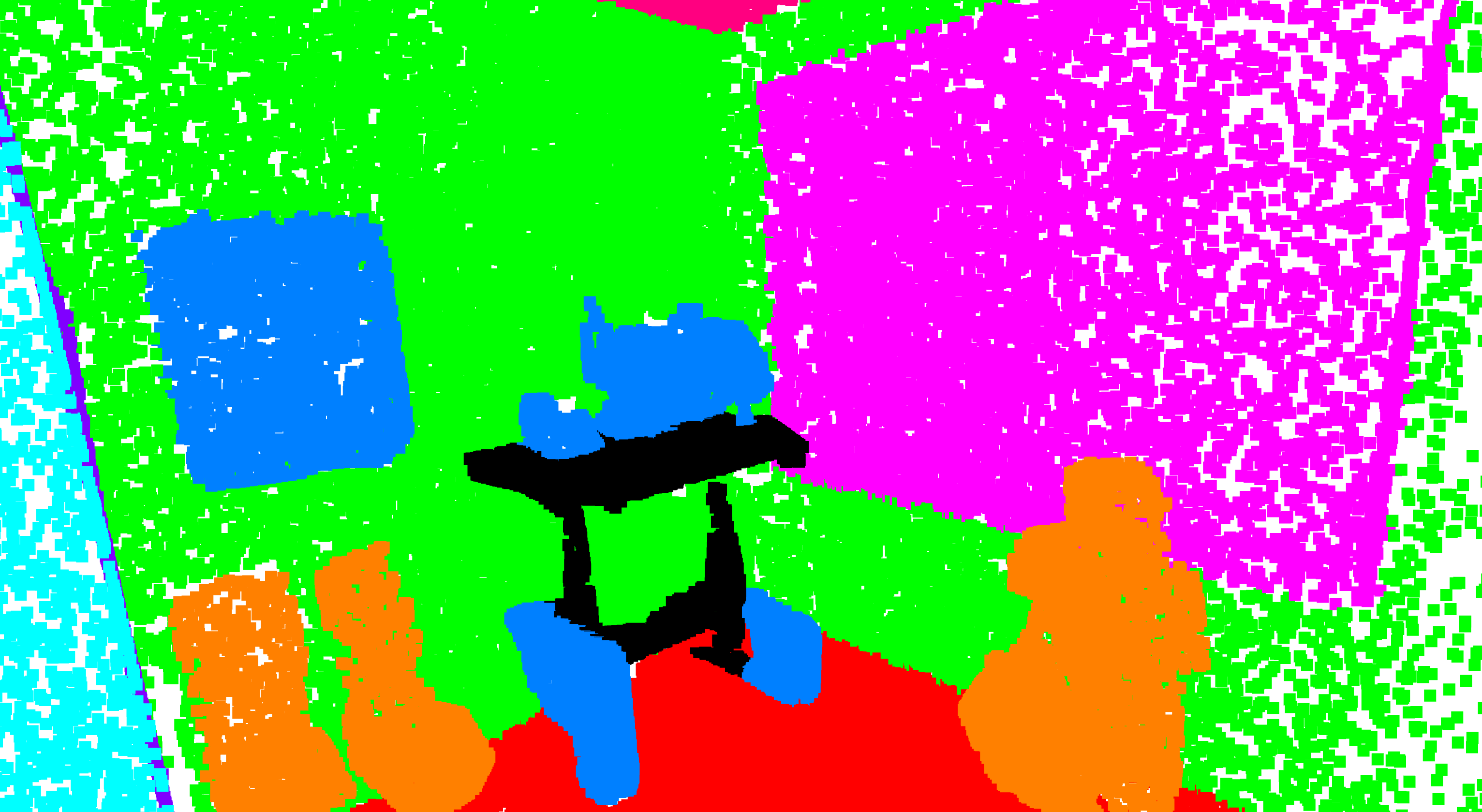}}\hspace{0.001cm}
\subfloat{\includegraphics[width=0.24\linewidth]{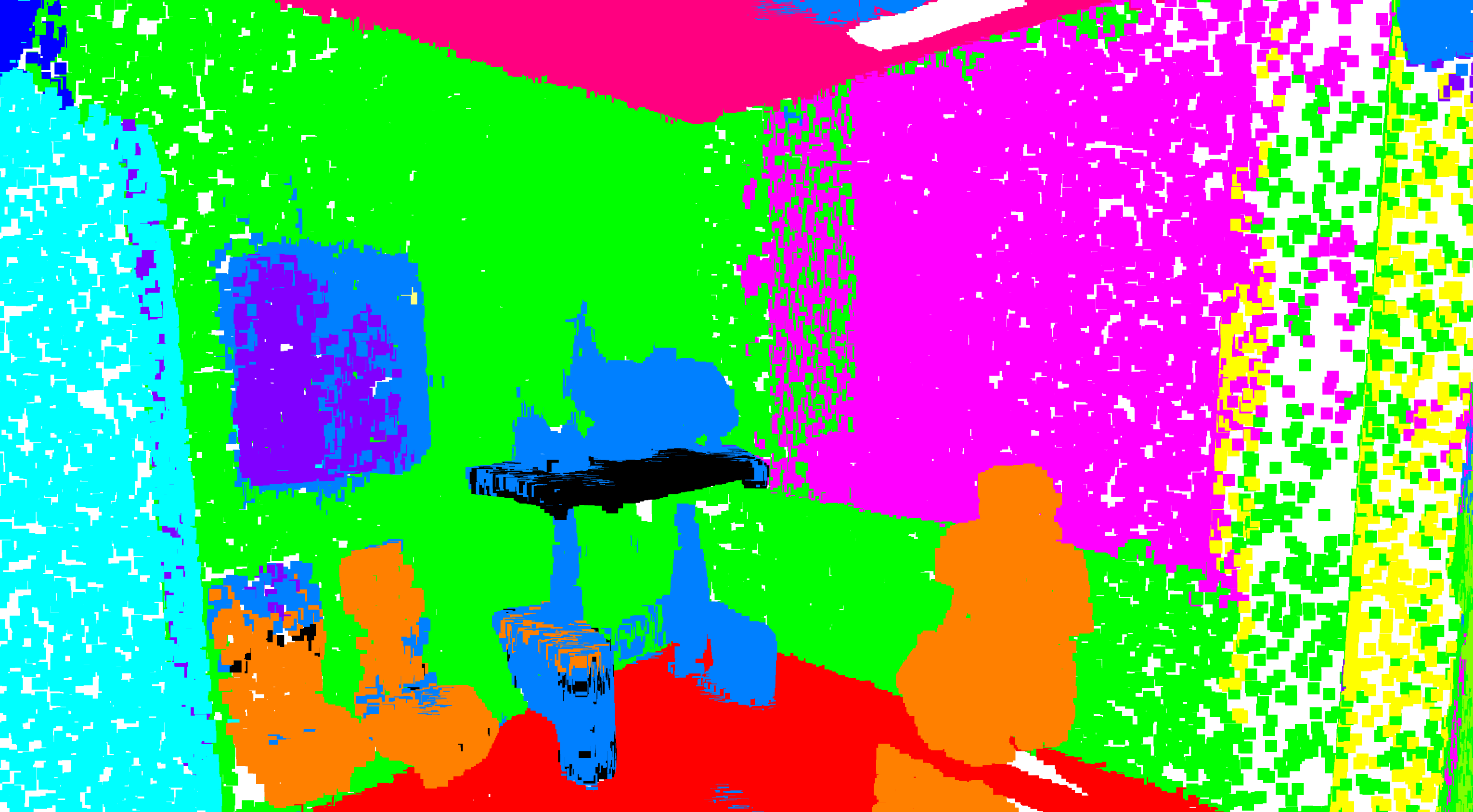}} \hspace{0.01cm}
% \vspace{0.1cm}

\subfloat{\includegraphics[width=0.24\linewidth]{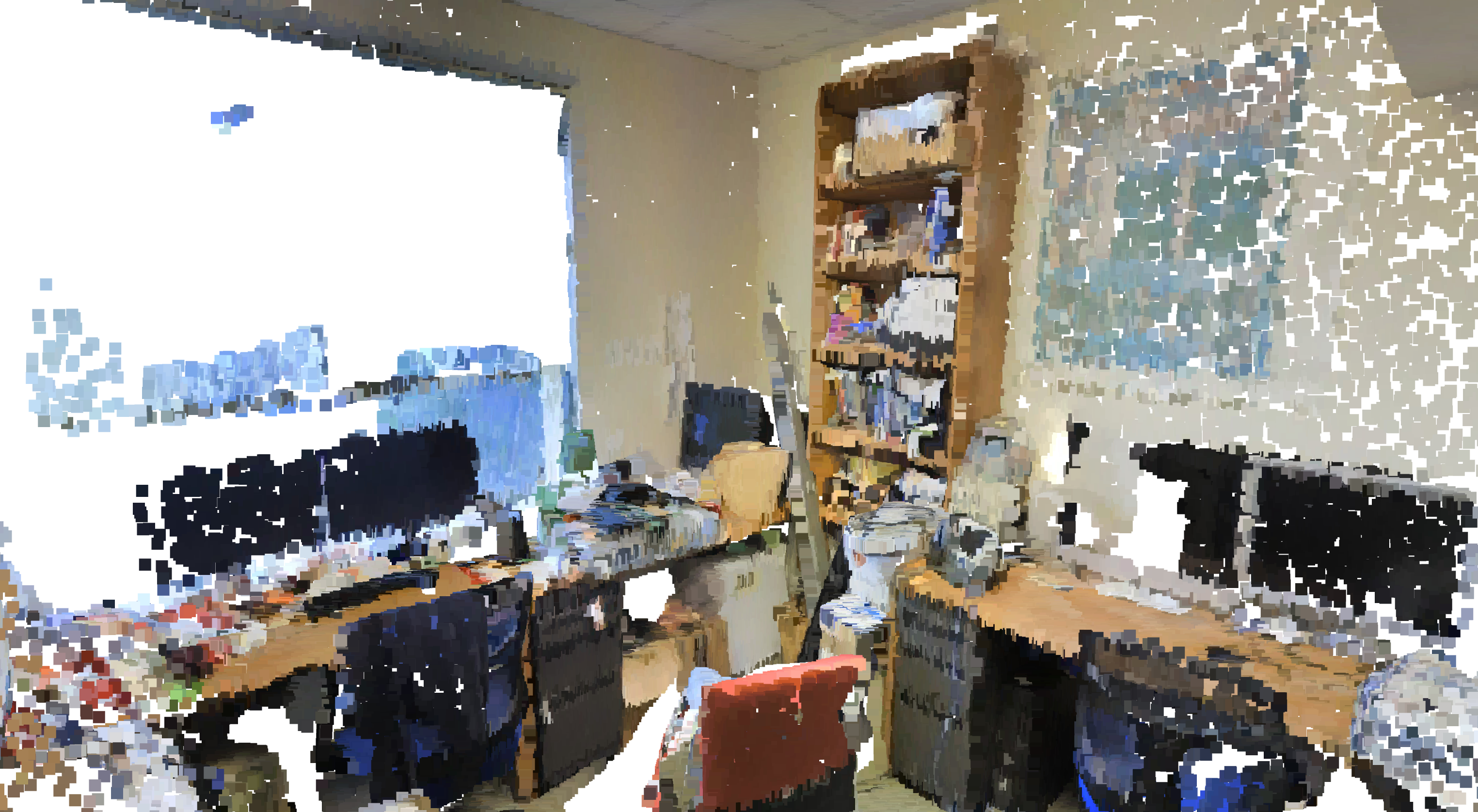}} \hspace{0.001cm}
\subfloat{\includegraphics[width=0.24\linewidth]{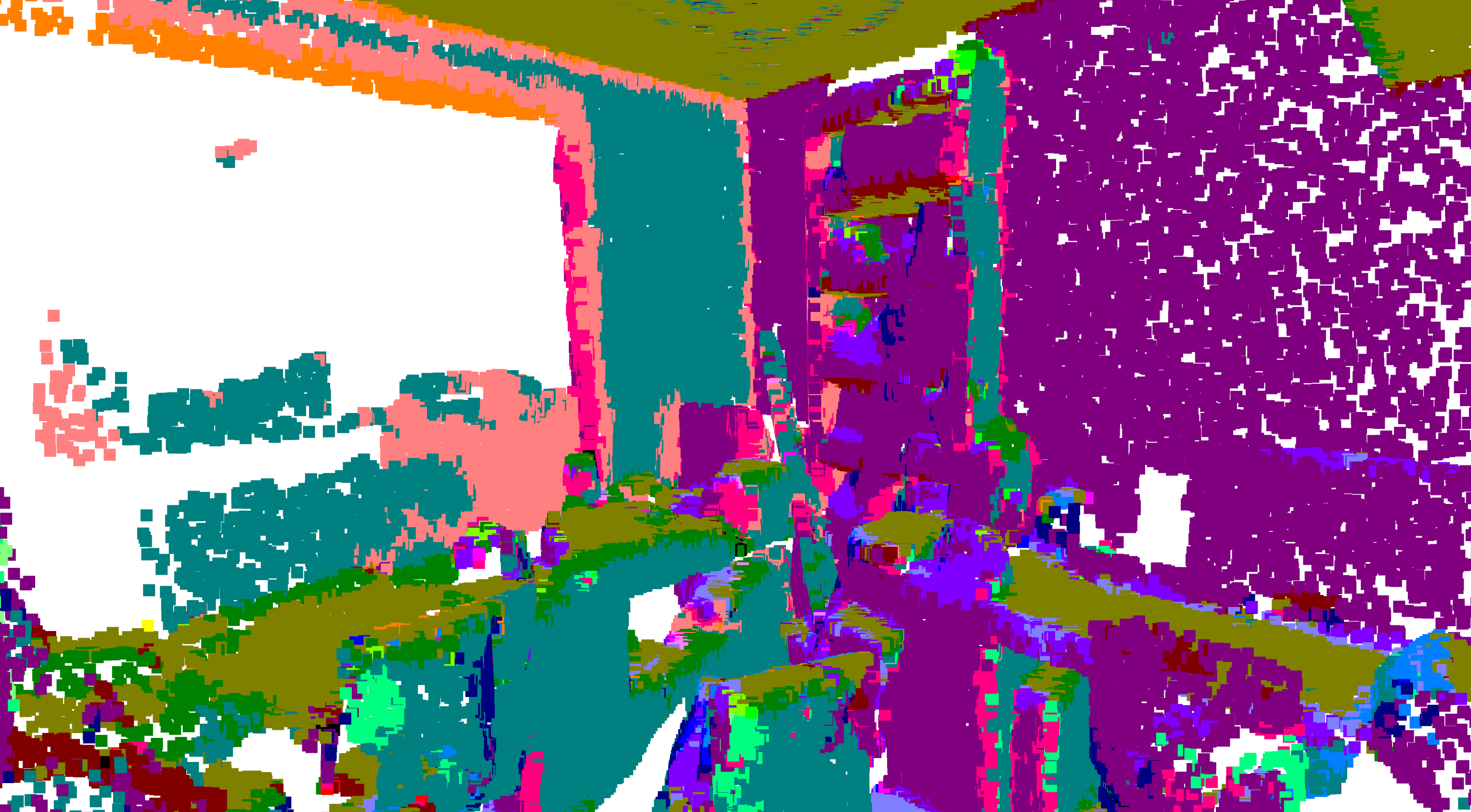}}\hspace{0.001cm}
\subfloat{\includegraphics[width=0.24\linewidth]{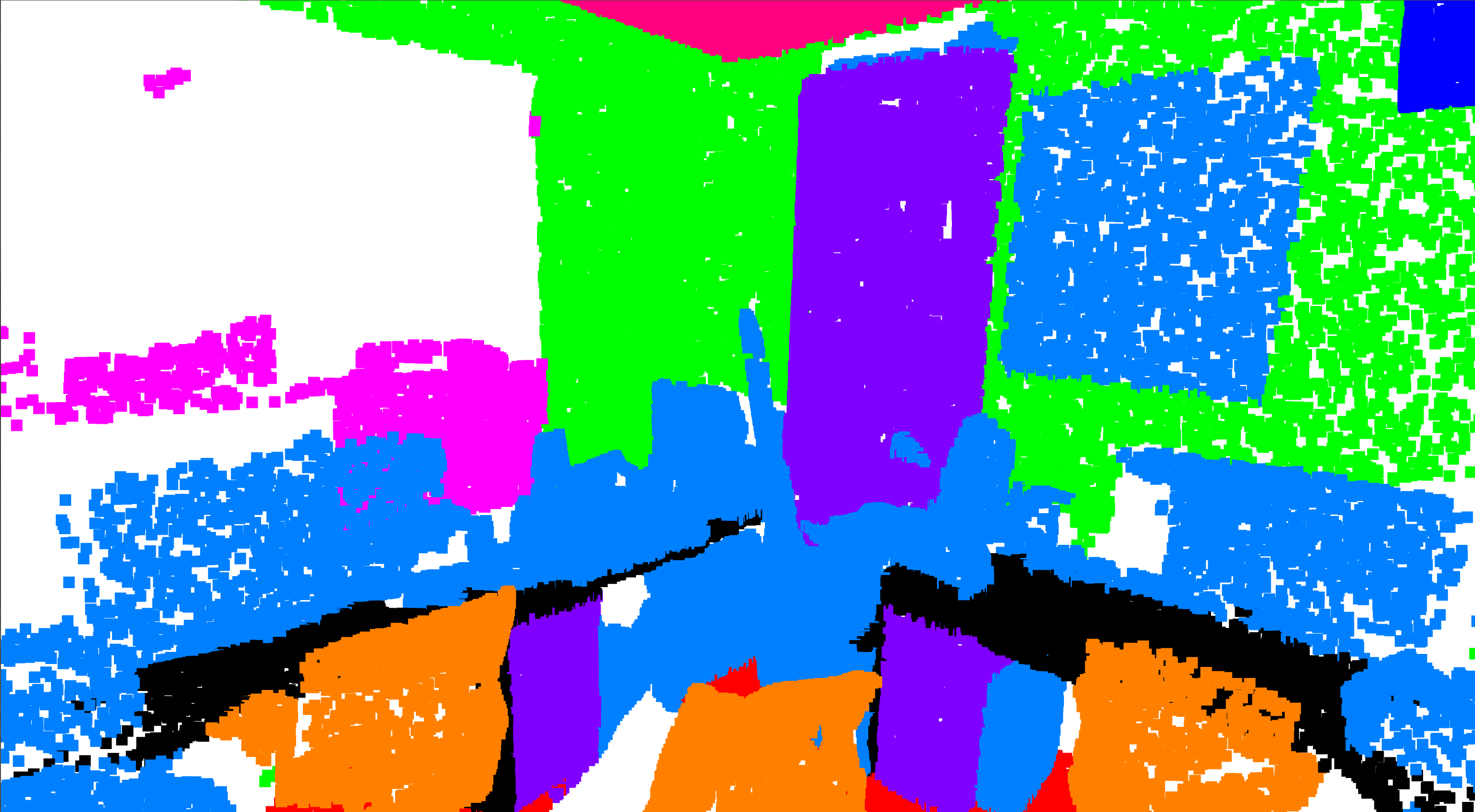}}\hspace{0.001cm}
\subfloat{\includegraphics[width=0.24\linewidth]{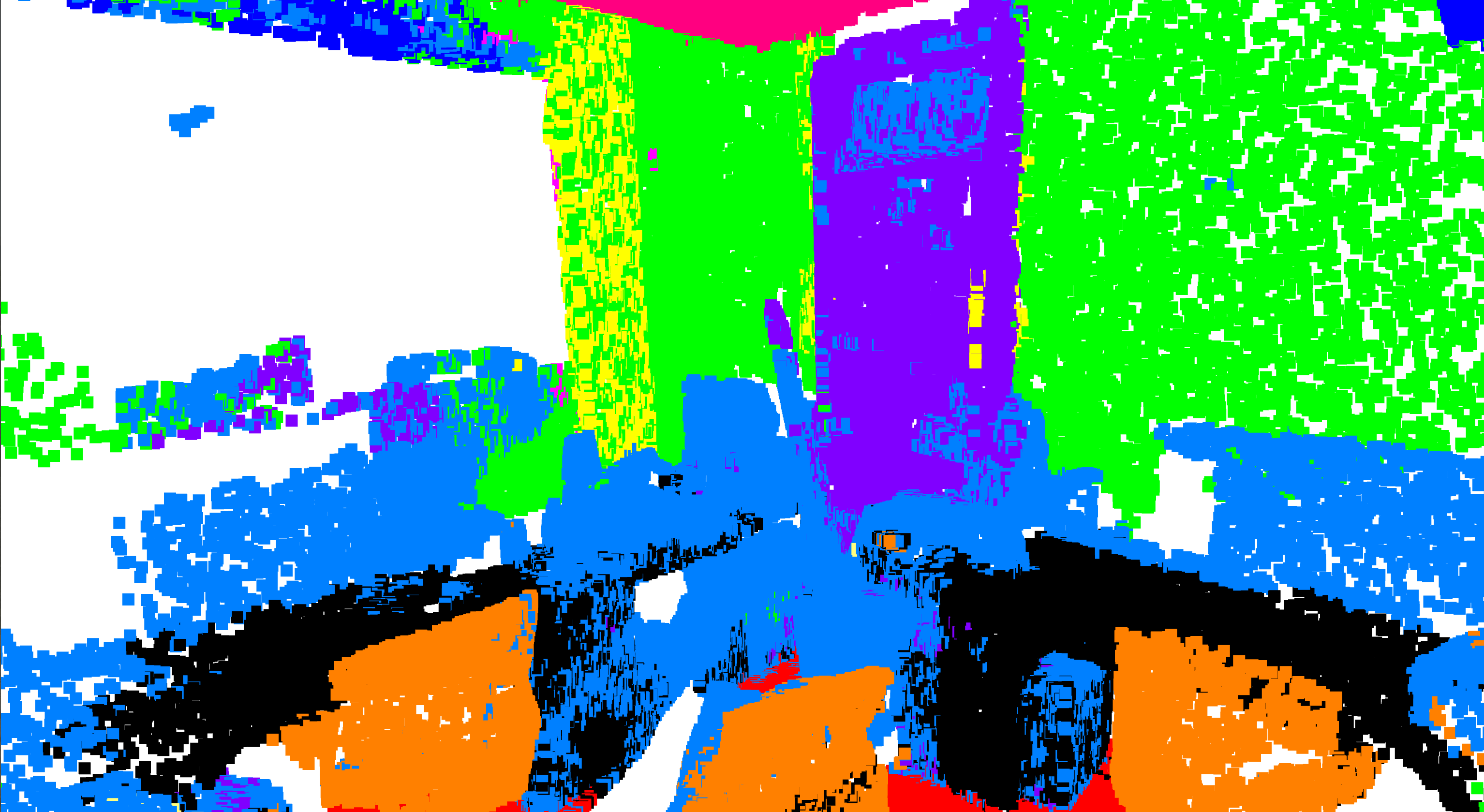}} \hspace{0.01cm}
% \vspace{0.1cm}

\subfloat{\includegraphics[width=0.24\linewidth]{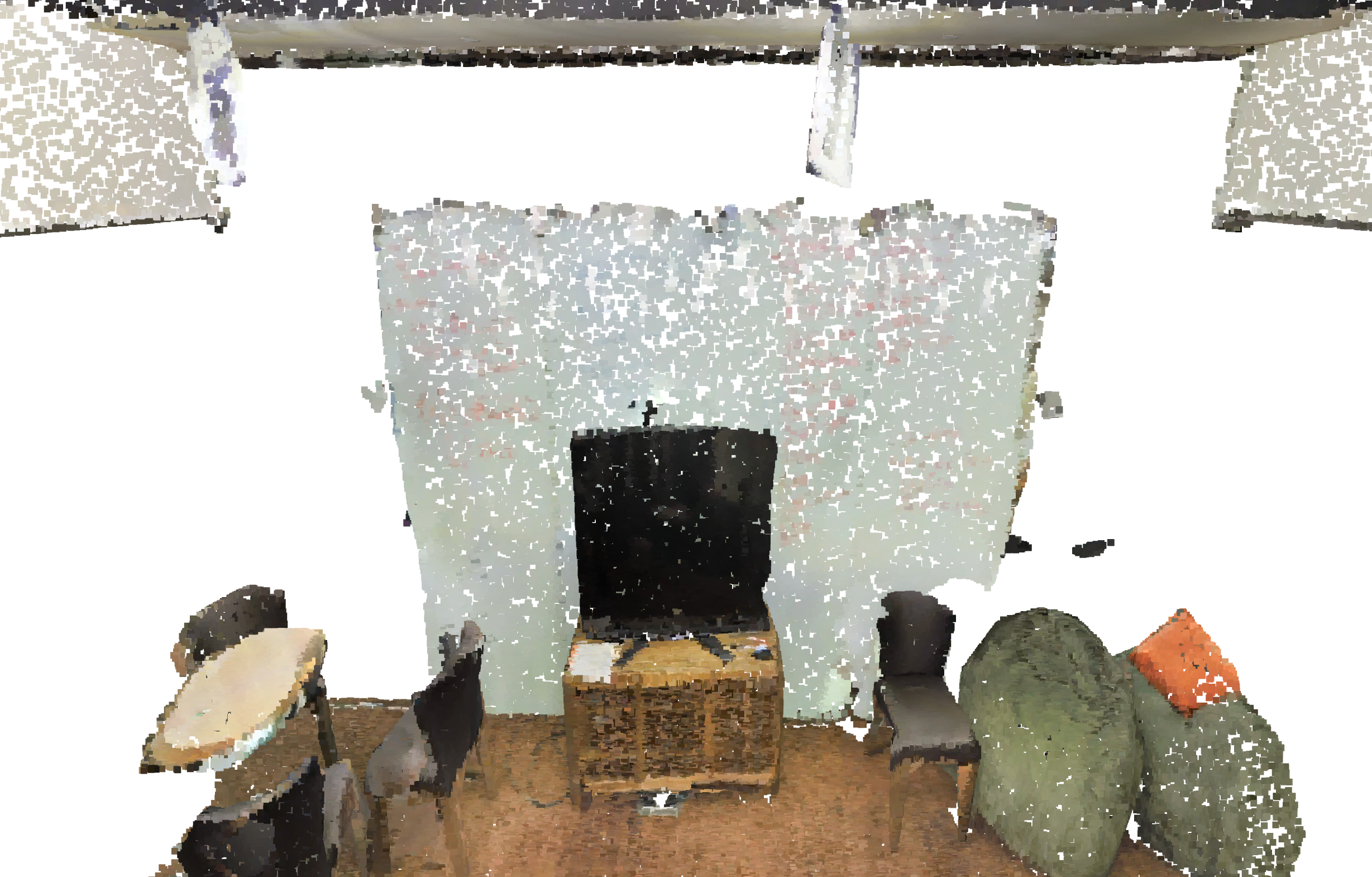}} \hspace{0.001cm}
\subfloat{\includegraphics[width=0.24\linewidth]{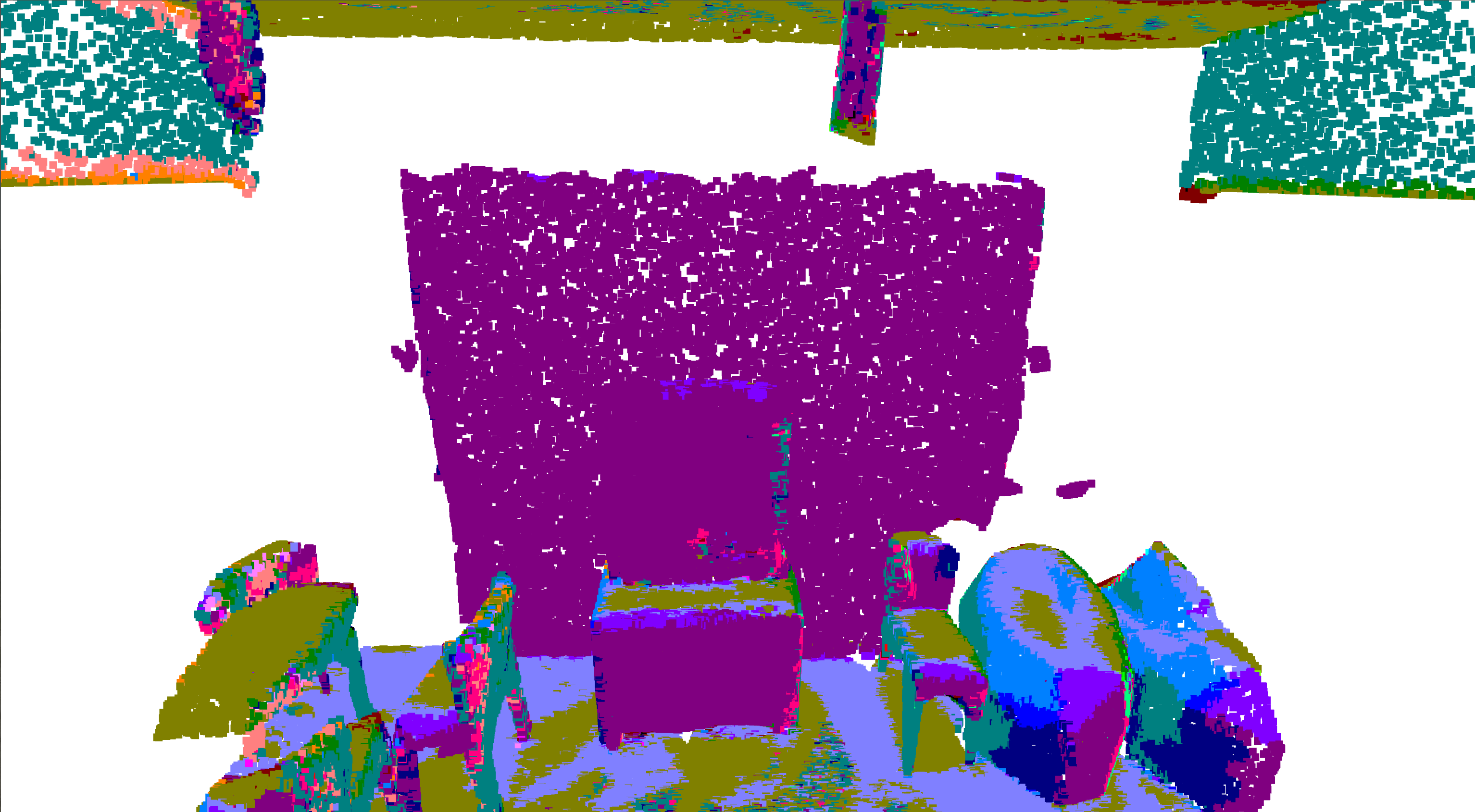}}\hspace{0.001cm}
\subfloat{\includegraphics[width=0.24\linewidth]{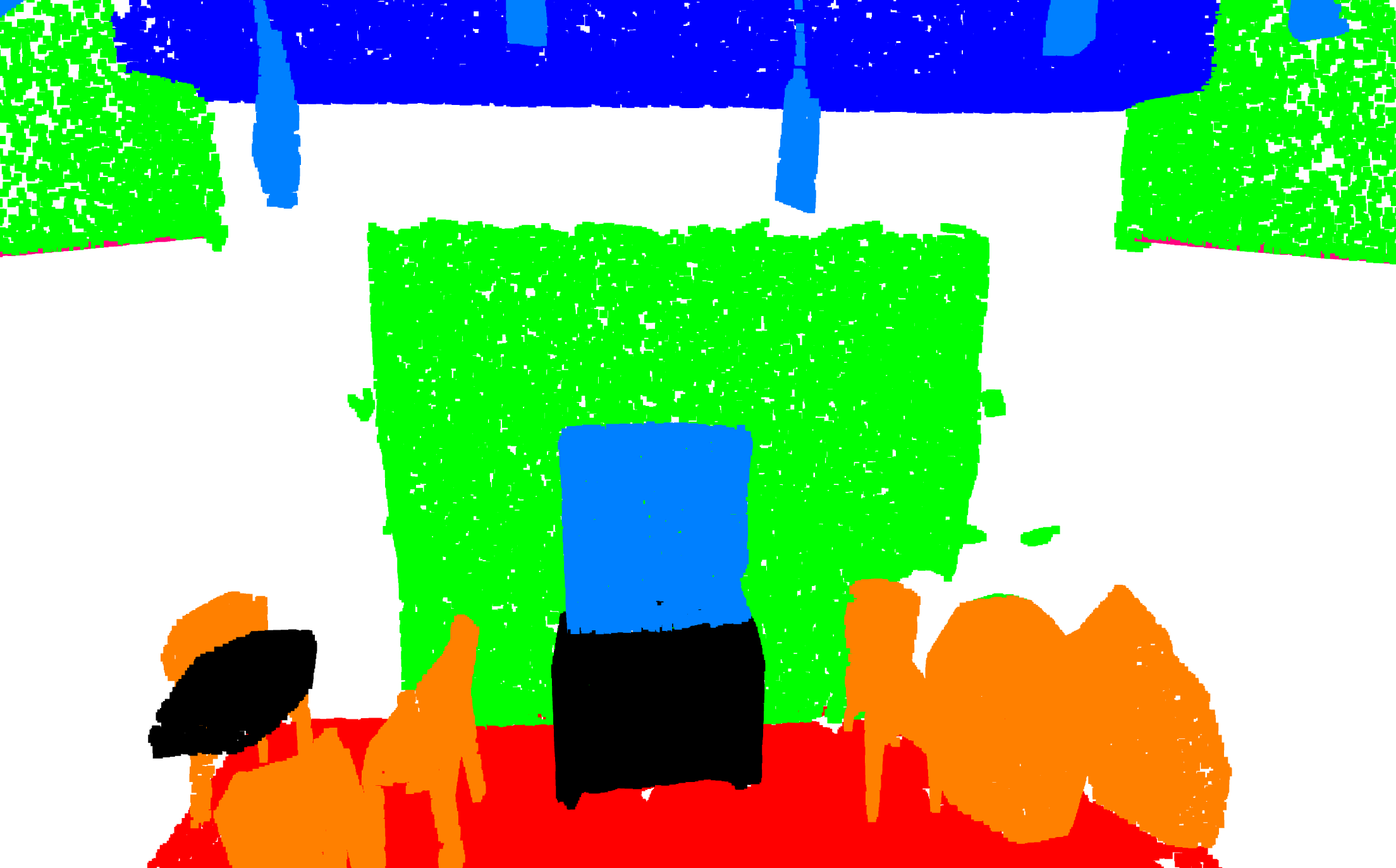}}\hspace{0.001cm}
\subfloat{\includegraphics[width=0.24\linewidth]{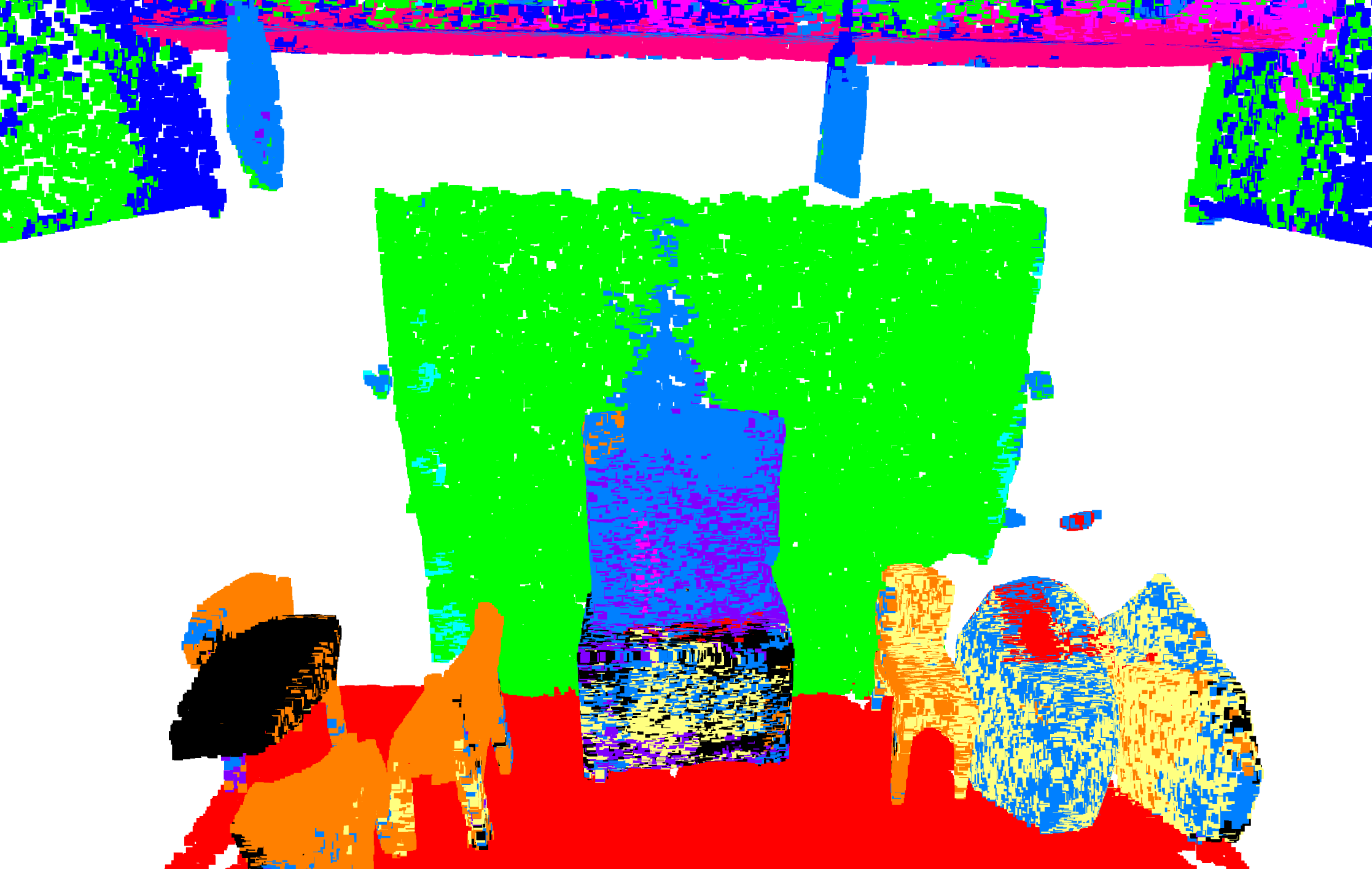}} \hspace{0.01cm}
% \vspace{0.1cm}

\subfloat{\includegraphics[width=0.24\linewidth]{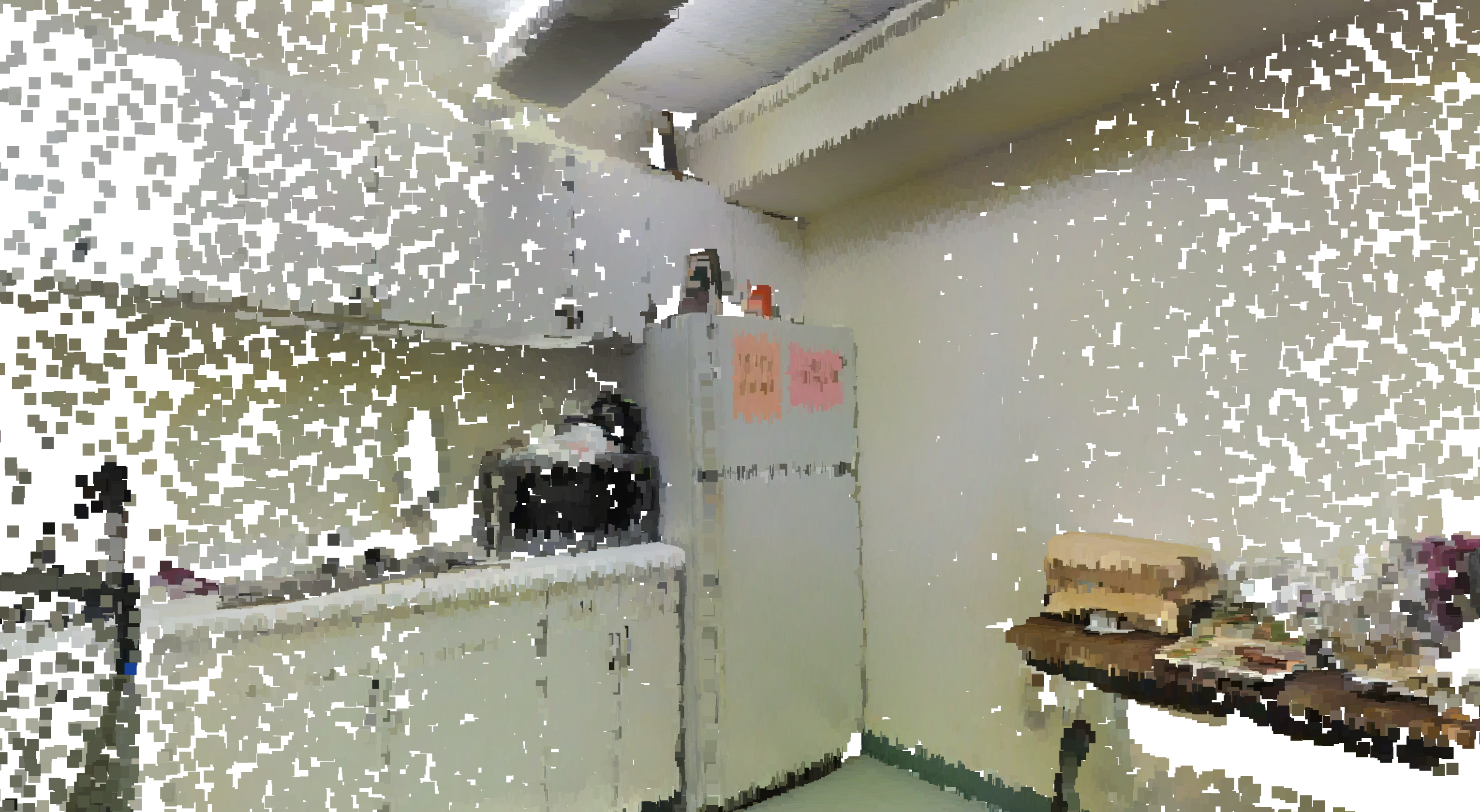}} \hspace{0.001cm}
\subfloat{\includegraphics[width=0.24\linewidth]{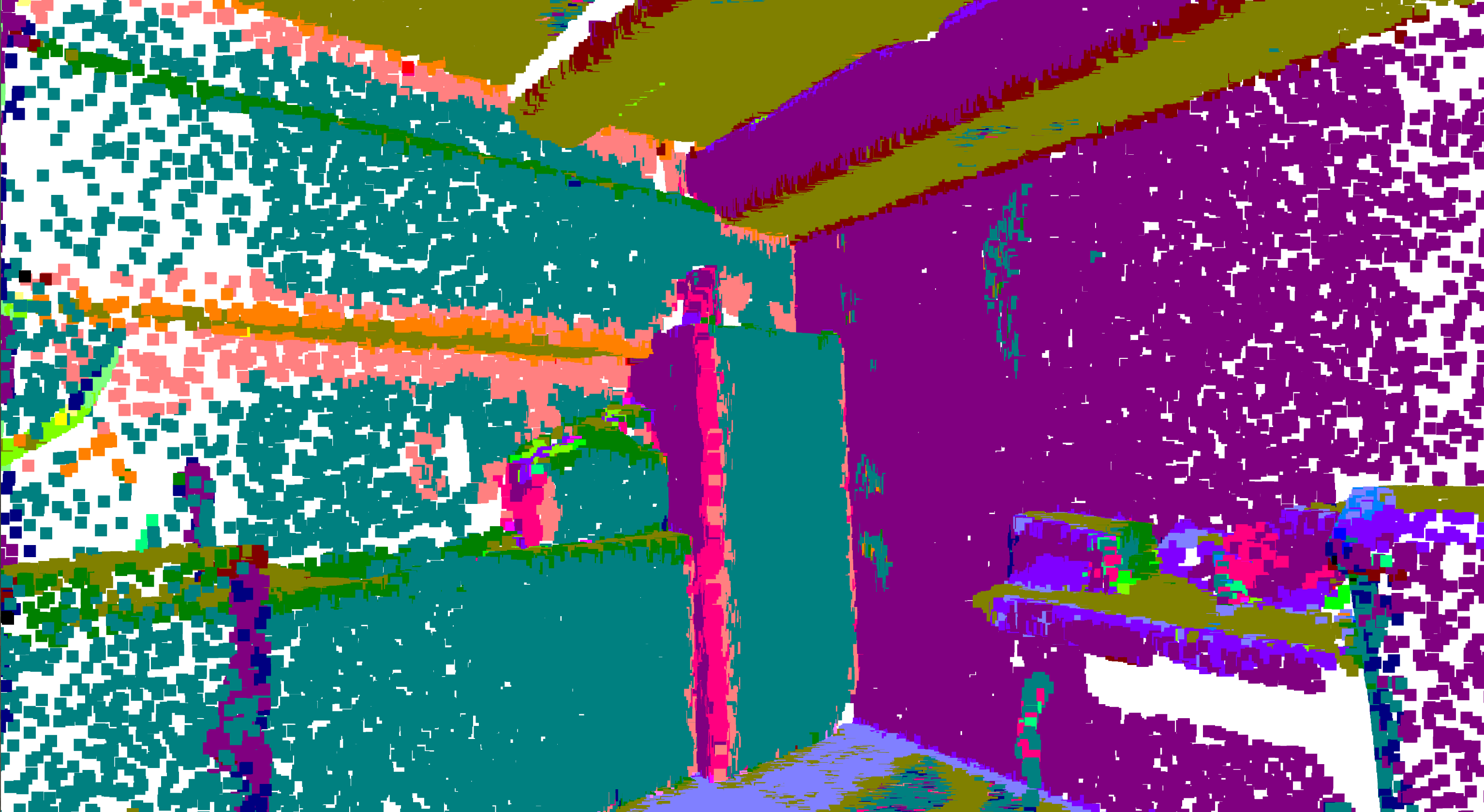}}\hspace{0.001cm}
\subfloat{\includegraphics[width=0.24\linewidth]{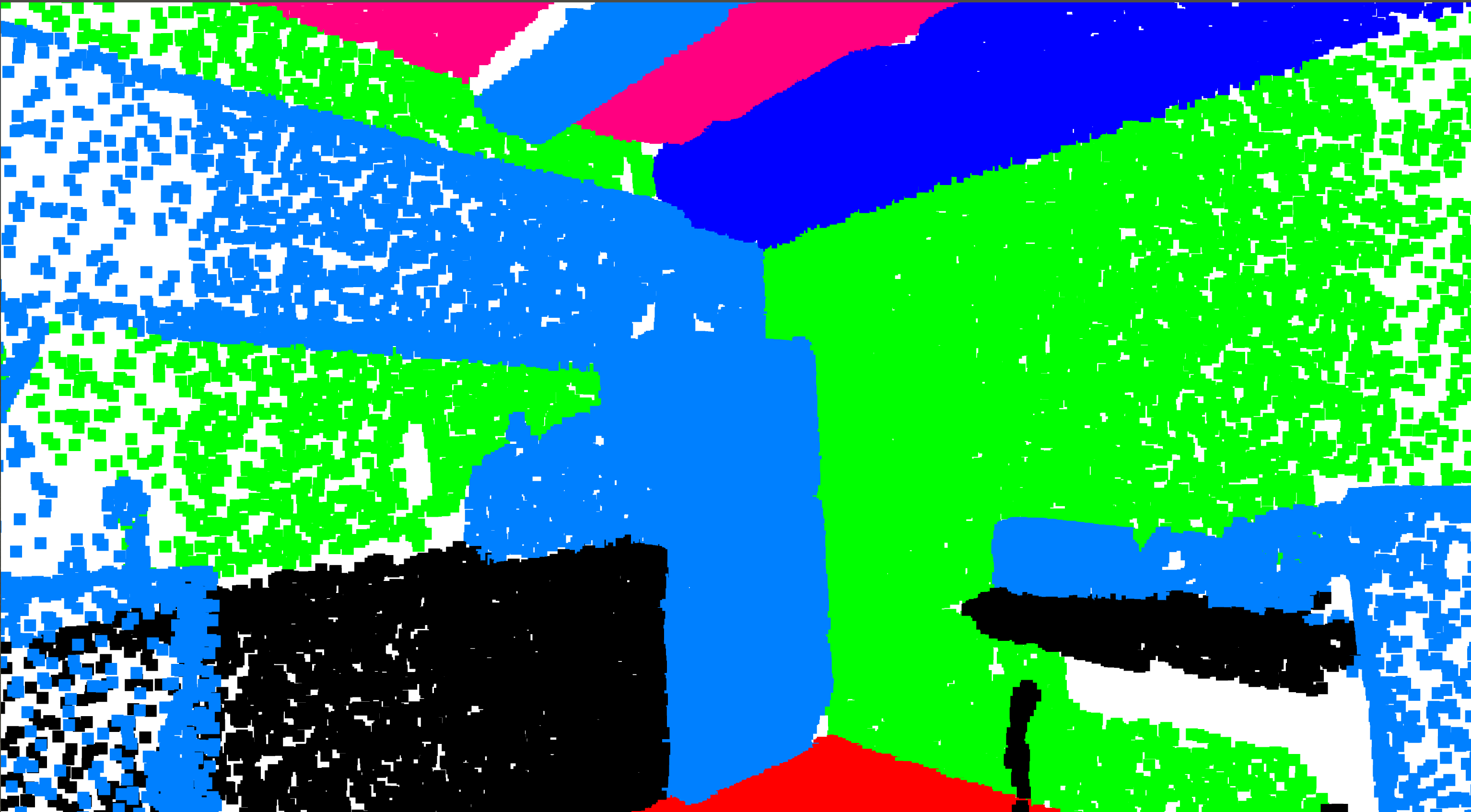}}\hspace{0.001cm}
\subfloat{\includegraphics[width=0.24\linewidth]{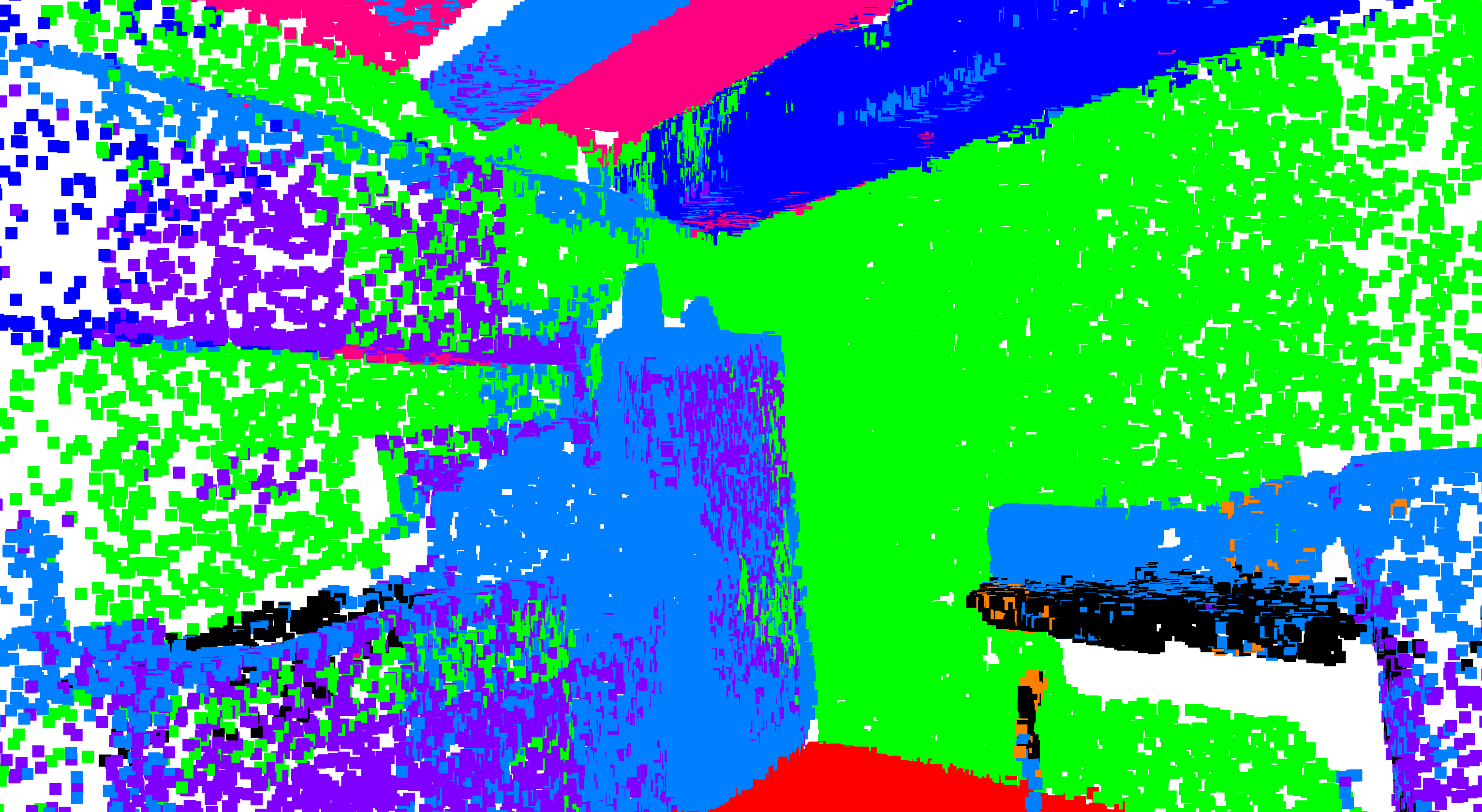}} \hspace{0.01cm}
\vspace{0.1cm}
\caption{Qualitative results for semantic segmentation, on S3DIS. [First column: input point clouds with color. Second column:  groups of points that approximately have the same surface normal, represented with the same color. Third column: ground -truth. Fourth column: proposed prediction.] }
\label{examples_s3dis}
\end{figure}

Second, the effect of attention modules has been assessed, with keeping the surface normals as extra channels ($c = 12$). Again, as shown in Table \ref{s3dis}, the best performance is achieved with a great increase in average per-class accuracy (that means a better accuracy for individual classes). The confusion matrix of test over area 6 can be seen in Figure \ref{fig:Confusion}, which is highly diagonal and shows good performance of the model. The actual qualitative results of the evaluation can be seen in Table \ref{examples_s3dis}.
The dataset was also split into 6 folds, and the k-fold results were extracted to be compared to other methods. The results are listed in Table \ref{kfold_s3dis}, where the proposed PAAConvNet obtains the best performance in some classes. In Fig. \ref{fig:modelsize}, it is shown that PAAConvNet displays competitive performance compared to other methods with a much lower number of parameters.

\subsection{Evaluation Results on SceneNN Dataset.}

The PAAConvNet was also trained and tested over SceneNN \cite{Hua2016SceneNNAS} dataset. The model has been trained by 56 scenes and tested on 20 other scenes. In this dataset, the same as S3DIS, the points close to each other are placed in groups of 4096 and each point belongs to one of 40 classes. 

\begin{figure}
\subfloat[RGB]{\hspace{2.2cm}}\subfloat[Normals]{\hspace{2.2cm}}\subfloat[GT]{\hspace{2.2cm}}\subfloat[Prediction]{\hspace{2.2cm}}
\vspace{0.1cm}
% \subfloat{\includegraphics[width=0.24\linewidth]{nno1.jpg}} \hspace{0.001cm}
% \subfloat{\includegraphics[width=0.24\linewidth]{nnn1.jpg}}\hspace{0.001cm}
% \subfloat{\includegraphics[width=0.24\linewidth]{nngt1.jpg}}\hspace{0.001cm}
% \subfloat{\includegraphics[width=0.24\linewidth]{nnp1.jpg}} \hspace{0.01cm}

\subfloat{\includegraphics[width=0.24\linewidth]{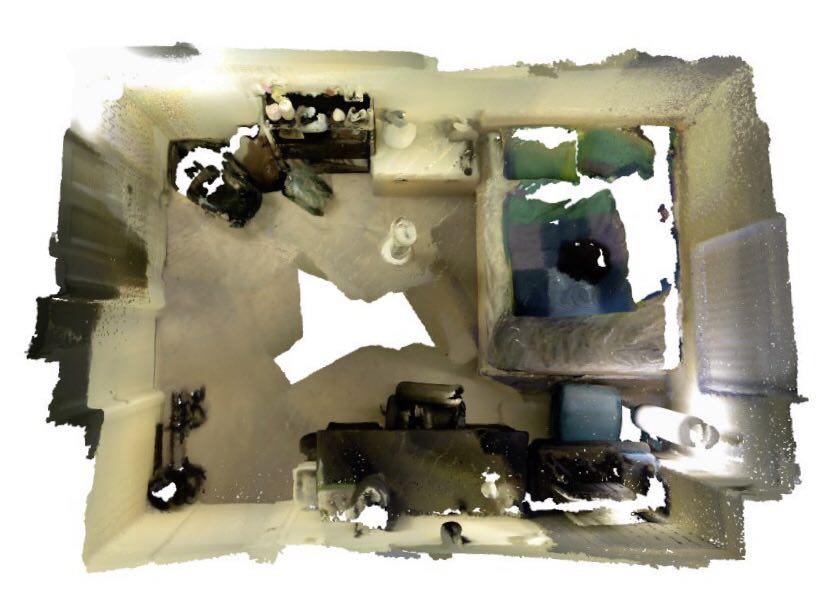}} \hspace{0.001cm}
\subfloat{\includegraphics[width=0.24\linewidth]{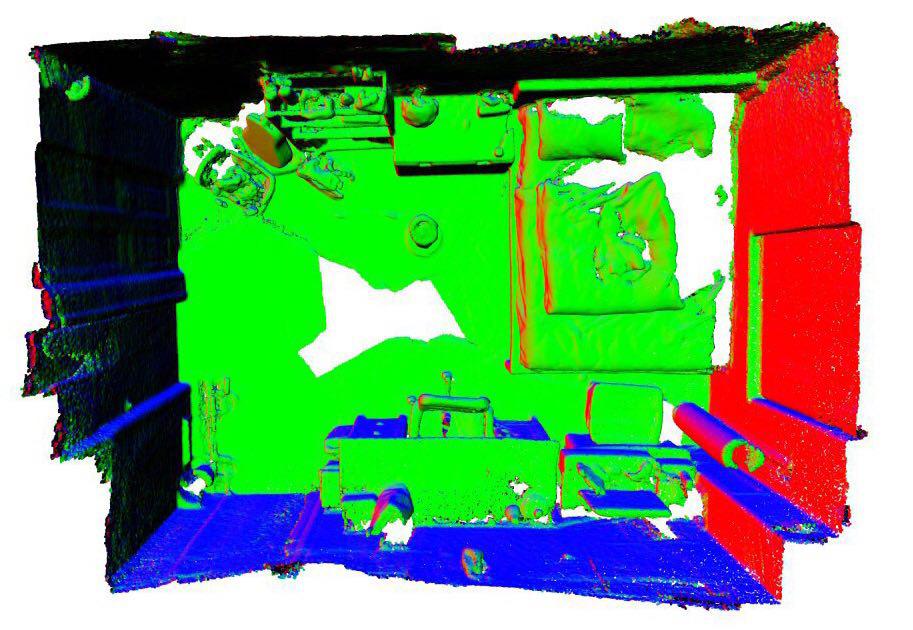}}\hspace{0.001cm}
\subfloat{\includegraphics[width=0.24\linewidth]{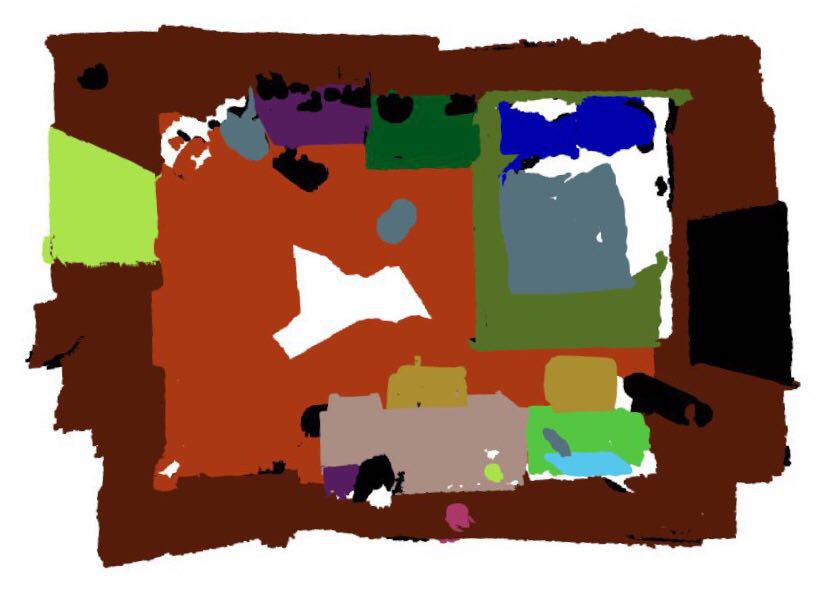}}\hspace{0.001cm}
\subfloat{\includegraphics[width=0.24\linewidth]{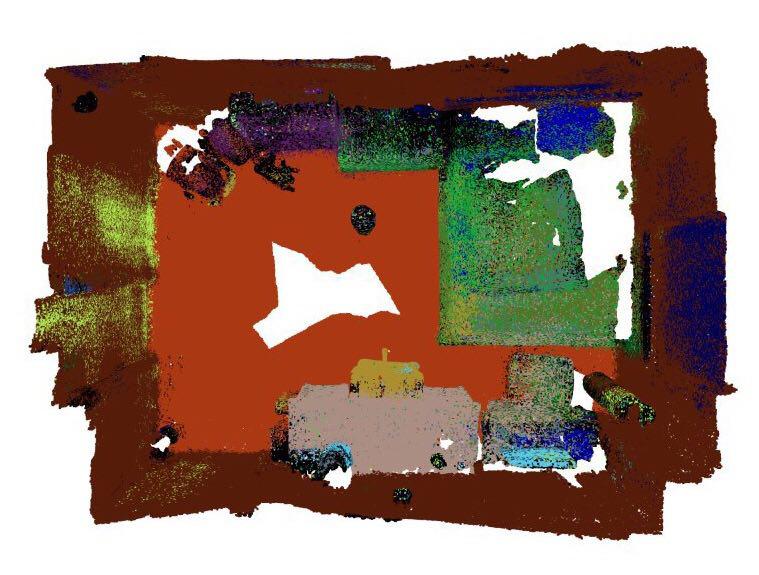}} \hspace{0.01cm}
% \vspace{0.1cm}

\subfloat{\includegraphics[width=0.24\linewidth]{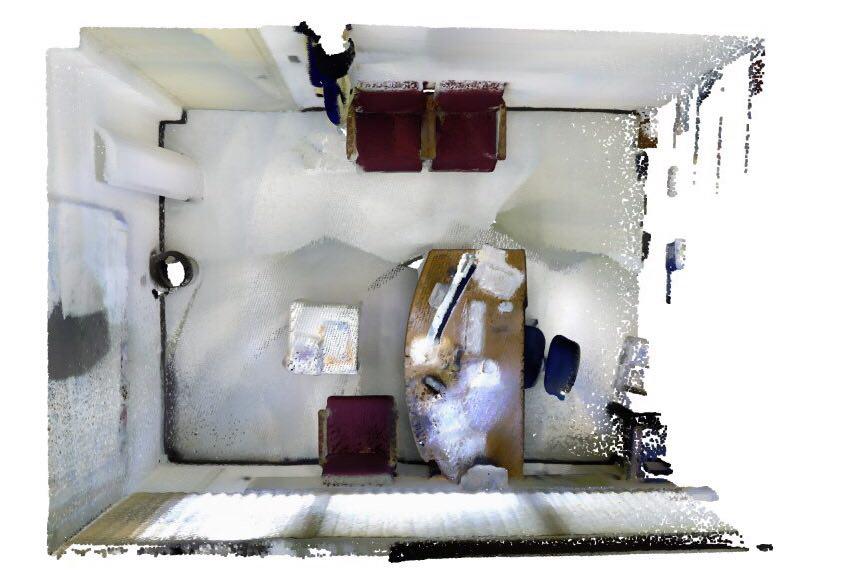}} \hspace{0.001cm}
\subfloat{\includegraphics[width=0.24\linewidth]{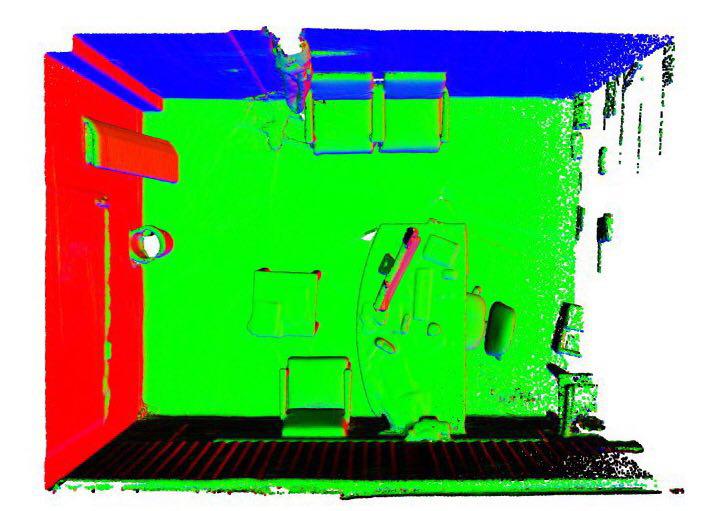}}\hspace{0.001cm}
\subfloat{\includegraphics[width=0.24\linewidth]{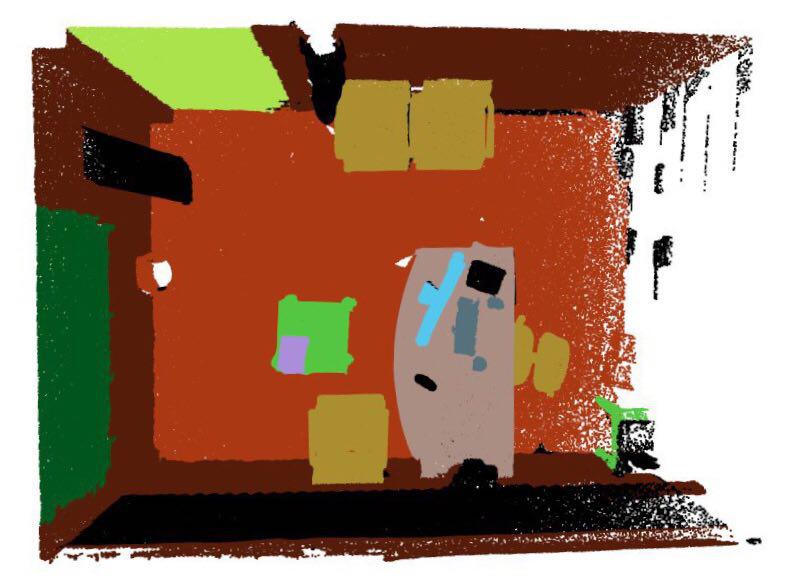}}\hspace{0.001cm}
\subfloat{\includegraphics[width=0.24\linewidth]{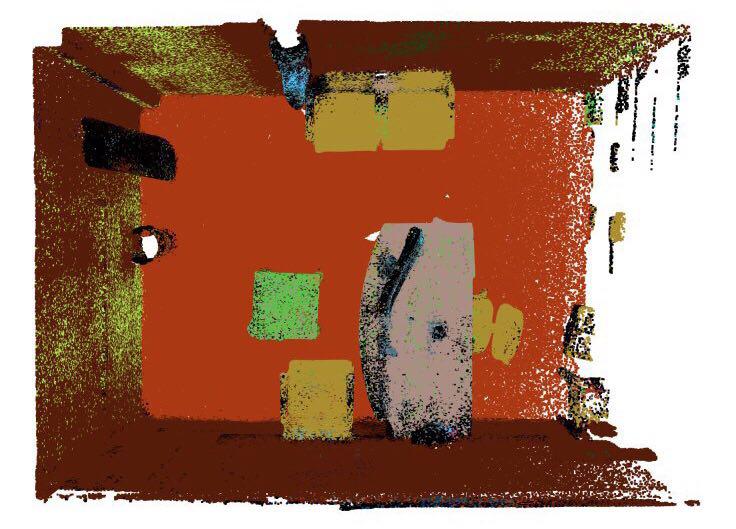}} \hspace{0.01cm}
% \vspace{0.1cm}

\subfloat{\includegraphics[width=0.24\linewidth]{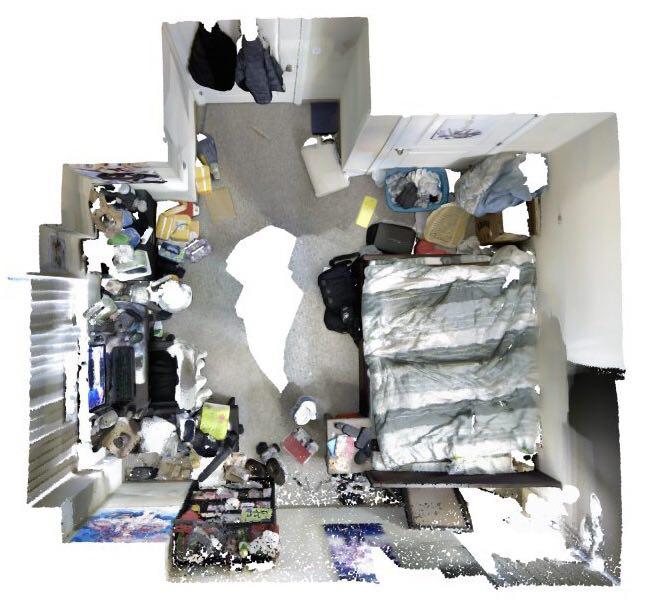}} \hspace{0.001cm}
\subfloat{\includegraphics[width=0.24\linewidth]{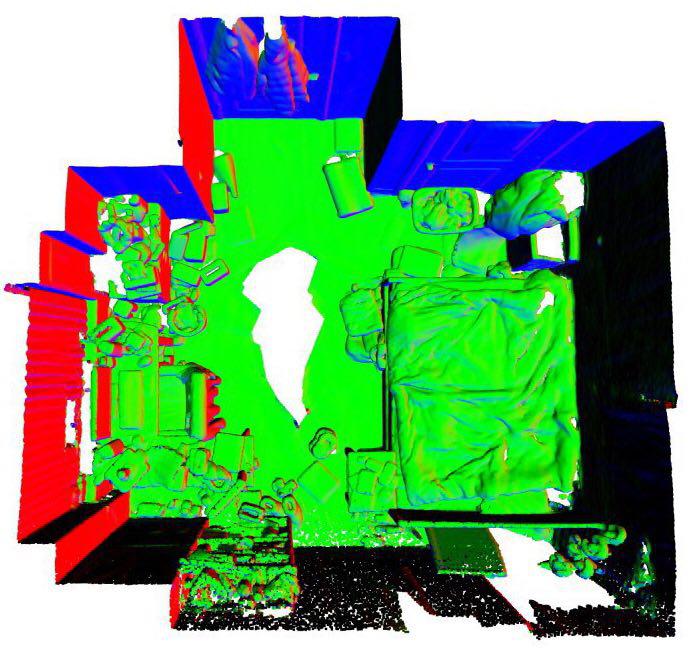}}\hspace{0.001cm}
\subfloat{\includegraphics[width=0.24\linewidth]{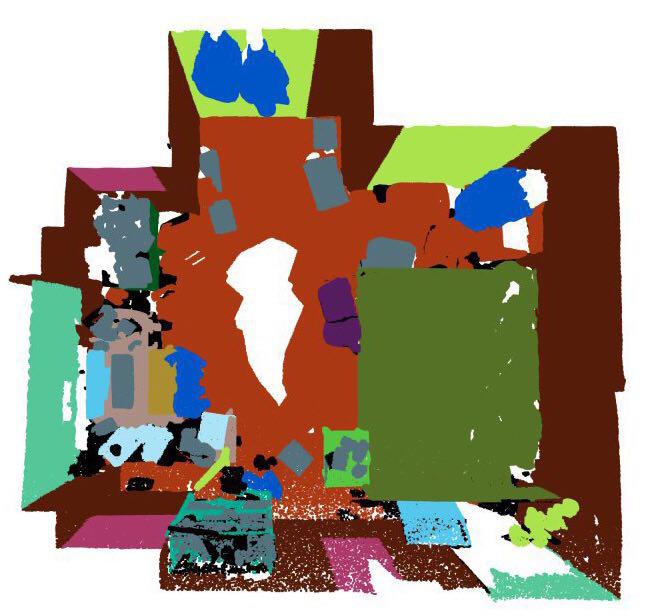}}\hspace{0.001cm}
\subfloat{\includegraphics[width=0.24\linewidth]{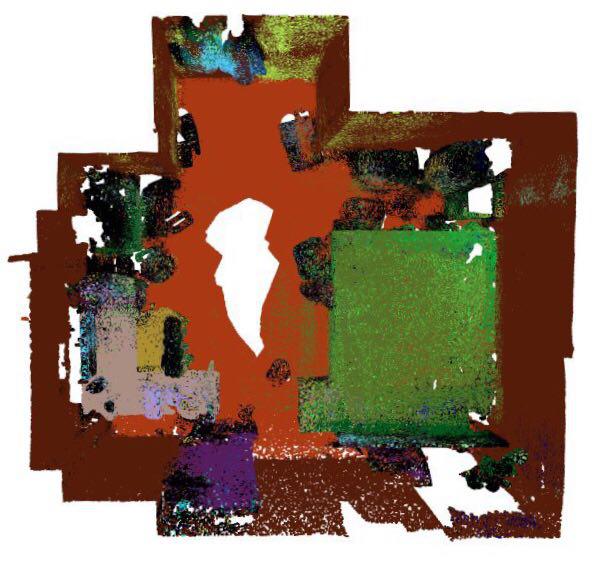}} \hspace{0.01cm}
% \vspace{0.1cm}
\caption{Qualitative results for semantic segmentation, on the SceneNN dataset. [First column: input point clouds with color. Second column: groups of points that approximately have the same surface normal, represented with the same color. Third column: ground-truth. Fourth column: proposed prediction.]}
\label{examples_NN}
\end{figure}

The per-class accuracy is reported in Table \ref{table:scenenn}, which shows the competitive results with the best performance in some classes compared to other methods and much improvement over the Pointwise. As it can be observed in qualitative results in Table \ref{examples_NN}, this dataset is much denser compared to S3DIS, thus requiring much more computations that means more testing and training time. Moreover, the number of classes is much higher, with some classes being under-represented in the dataset. These factors generally contribute to the lower accuracy achieved on this dataset. 

\section{Conclusion}
\label{sec:5}
The attention-based convolutional neural network model was proposed for 3D semantic segmentation. The model considered 3 main challenges of 3D point cloud labeling with the corresponding solutions to outperform the previous approaches in terms of accuracy and model size. To consider 3D points contextual information in unordered irregular sets, the atrous pointwise convolution method was proposed with the spatial attention modules to highlight the more important point sets. The performance of the proposed method was compared with the state-of-the-art models. It obtained better accuracy with a fewer number of parameters than those methods on standard benchmarks.


% \section*{References}
\bibliographystyle{IEEEtran}
\bibliography{3dss_ref}

\end{document}